\documentclass{article}
\usepackage{sahoo_diffusion_new}

\usepackage{amsmath}
\usepackage{amssymb}
\usepackage{amsfonts}
\usepackage{amsthm}
\usepackage{mathtools}
\usepackage{bm}
\usepackage{bbm}
\usepackage{cases}
\usepackage{dsfont}
\usepackage{cases}
\usepackage{nicefrac}
\usepackage{MnSymbol}
\usepackage{bbding}
\usepackage{pifont}

\usepackage[utf8]{inputenc} %
\usepackage[T1]{fontenc}    %
\usepackage{textcomp}

\usepackage{graphicx}
\usepackage{wrapfig}
\usepackage{float}
\usepackage{caption}
\usepackage{subcaption}     %
\usepackage{array}
\usepackage{multirow}
\usepackage{multicol}
\usepackage{booktabs}
\usepackage{makecell}
\usepackage{colortbl}

\usepackage{xr-hyper}
\usepackage{hyperref}       %
\usepackage[capitalize,noabbrev]{cleveref}
\usepackage{url}

\usepackage[makeroom]{cancel}

\hypersetup{
  final,
  colorlinks,
  linkcolor=ourred,
  citecolor=ourblue,
  urlcolor=url,
}

\usepackage{algorithm}
\usepackage{algpseudocode}

\usepackage[page,toc,titletoc,title]{appendix}
\usepackage{xargs}
\usepackage{xspace}
\usepackage[normalem]{ulem}
\usepackage{soul}
\usepackage{fancyvrb}

\usepackage{lipsum}
\usepackage{blindtext}

\usepackage{xcolor}
\definecolor{ourblue}{rgb}{0.368,0.507,0.71}
\definecolor{ourorange}{rgb}{0.881,0.611,0.142}
\definecolor{ourgreen}{rgb}{0.56,0.692,0.195}
\definecolor{ourgreen2}{rgb}{0.46,0.792,0.195}
\definecolor{ourred}{rgb}{0.923,0.386,0.209}
\definecolor{ourviolet}{rgb}{0.528,0.471,0.701}
\definecolor{ourbrown}{rgb}{0.772,0.432,0.102}
\definecolor{ourlightblue}{rgb}{0.364,0.619,0.782}
\definecolor{ourbrightblue}{RGB}{93,135,237}
\definecolor{ourdarkgreen}{rgb}{0.572,0.586,0.}

\definecolor{ourcyan2}{rgb}{0.125,0.722,0.804}
\definecolor{ourred2}{rgb}{0.863,0.184,0.047}
\definecolor{ouryellow2}{cmyk}{0,0.16,1.0,0.07}
\definecolor{ourviolet2}{cmyk}{0.55,0.56,0,0.47}
\definecolor{ourorange2}{cmyk}{0,0.46,0.89,0.11}
\definecolor{grayseq}{RGB}{120,120,120}

\definecolor{discretecolor}{RGB}{11,83,150}
\definecolor{gaussiancolor}{RGB}{230,145,56}
\definecolor{argmaxcolor}{RGB}{154,0,0}

\definecolor{customred}{RGB}{169, 45, 59}
\definecolor{url}{HTML}{d95225}

\definecolor{olivegreen}{RGB}{165,185,115}
\definecolor{olivegreendark}{RGB}{145,165,95}

\definecolor{maroon}{RGB}{140, 45, 45}

\usepackage{soul}

\newcommand{\Fig}[1]{Figure~\ref{#1}}  %
\newcommand{\fig}[1]{Fig.~\ref{#1}}    %

\newcommand{\tab}[1]{Table~\ref{#1}}
\newcommand{\alg}[1]{Algo. \ref{#1}}
\newcommand{\Eqn}[1]{(\ref{#1})}
\renewcommand{\sec}[1]{Sec.~\ref{#1}} %
\newcommand{\supp}[1]{Suppl.~\ref{#1}}

\makeatletter
\DeclareRobustCommand\onedot{\futurelet\@let@token\@onedot}
\def\@onedot{\ifx\@let@token.\else.\null\fi\xspace}

\makeatother

\newcommand{\mask}{[\textsc{mask}]~}

\newtcolorbox{remarkbox}{
  colback=blue!12,
  colframe=blue!70!black,
  left=3.5pt,
  right=3.5pt,
  top=3pt,
  bottom=2pt
}

\newtcolorbox{corollarybox}{
  colback=olivegreen!12,
  colframe=olivegreen!70!black,
  left=3.5pt,
  right=3.5pt,
  top=3pt,
  bottom=2pt
}

\newtcolorbox{limitationbox}{
  colback=red!12,
  colframe=red!40!black,
  left=3.5pt,
  right=3.5pt,
  top=3pt,
  bottom=2pt
}

\usepackage{amsmath,amsfonts,bm}

\def\eqref#1{equation~\ref{#1}}

\def\1{\bm{1}}

\def\vone{{\bm{1}}}

\DeclareMathAlphabet{\mathsfit}{\encodingdefault}{\sfdefault}{m}{sl}
\SetMathAlphabet{\mathsfit}{bold}{\encodingdefault}{\sfdefault}{bx}{n}

\def\gD{{\mathcal{D}}}

\def\gL{{\mathcal{L}}}

\newcommand{\E}{\mathbb{E}}

\DeclareMathOperator*{\argmax}{arg\,max}

\def\x{{\mathbf x}}
\def\z{{\mathbf z}}
\def\c{{\mathbf c}}

\def\kl{\text{D}_{\text{KL}}}

\def\cat{\text{Cat}}
\def\x{{\mathbf x}}

\def\xi{{\mathbf x}^{(i)}}
\def\c{{\mathbf c}}

\def\c{{\mathbf c}}

\def\m{{\mathbf m}}

\def\pst{{p^\theta_{s|t}}}

\def\kl{\text{D}_{\text{KL}}}

\def\studentw{\bm \theta}
\def\teacherw{\bm \theta_0}

\def\at{{\alpha_{t}}}

\def\dat{{\alpha'_{t}}}

\def\as{{\color{discretecolor} \alpha_{s}}}
\def\denoise{\mathbf {x}_\theta}

\def\student{\x_{\studentw}}
\def\teacher{\x_{\teacherw}}

\def\felbo{\textit{f}}

\def\prior{\text{$\boldsymbol{\mathit{\pi}}$}}

\def\c{{\mathbf c}}

\newcommand{\onehotset}{\mathcal{V}}
\newcommand{\barx}{\bar{\mathbf{x}}}

\def\supL{{\color{grayseq}1:L}}
\def\supl{{\ell}}
\def\supi{{\color{grayseq}i}}

\def\ztL{\z_t^{\supL}}
\def\ztl{\z_t^{\supl}}

\def\xl{\x^{\supl}}
\def\xll{\x^{<\supl}}

\def\xi{\x^{\supi}}

\def\pst{p^{\theta}_{s|t}}
\def\na{{\color{grayseq}\text{N/A}}}

\usepackage{fancyvrb}
\usepackage{pythonhighlight}

\theoremstyle{plain}

\theoremstyle{definition}

\theoremstyle{remark}

\tcbuselibrary{listings,skins}
\usepackage{listings}
\usepackage{multirow}
\usepackage{array}
\usepackage{graphicx} 
\usepackage{booktabs}
\usepackage{url}
\usepackage{hyperref}
\usepackage{bbm} %
\usepackage{pifont} %
\usepackage{autonum}

\usepackage{subcaption}
\usepackage{xspace}

\usepackage{rotating}

\usepackage{amsthm}
\usepackage{wrapfig}

\usepackage[table]{xcolor}
\usepackage[most]{tcolorbox}

\definecolor{zaibluebg}{HTML}{E6F2FF}
\definecolor{zaibluefg}{HTML}{007AFF}
\newcolumntype{Z}{>{\columncolor{zaibluebg}\color{zaibluefg}}c}
\newcommand{\normalZ}[1]{\multicolumn{1}{c}{{\color{black}#1}}}

\definecolor{newnewourred}{RGB}{250,190,190}

\newcommand{\newnewourredcell}[1]{%
  \cellcolor{newnewourred}{\color{black}#1}%
}
\newcommand{\newnewourredcellgrad}[2]{%
  \cellcolor{newnewourred!#1!white}{\color{black}#2}%
}
\DeclareRobustCommand{\newnewourredsample}[1]{%
  \begingroup
  \setlength{\fboxsep}{1pt}%
  \colorbox{newnewourred}{\color{black}\strut #1}%
  \endgroup
}

\tcbset{
  generationbase/.style={
    enhanced,
    listing only,
    listing engine=listings,
    fonttitle=\bfseries\small,
    coltitle=black,
    boxrule=0.5pt,
    arc=1pt,
    left=1mm,
    right=1mm,
    top=1mm,
    bottom=1mm,
    listing options={
      basicstyle=\ttfamily\scriptsize,
      columns=fullflexible,
      keepspaces=true,
      breaklines=true,
      showstringspaces=false
    }
  }
}

\newtcblisting{flaggedgeneration}[1]{
  generationbase,
  title={#1},
  colback=red!2,
  colframe=red!55!black,
  colbacktitle=red!10
}

\newtcblisting{nearmissgeneration}[1]{
  generationbase,
  title={#1},
  colback=orange!3,
  colframe=orange!65!black,
  colbacktitle=orange!12
}

\newtcblisting{cleangeneration}[1]{
  generationbase,
  title={#1},
  colback=green!2,
  colframe=green!45!black,
  colbacktitle=green!10
}

\newcolumntype{q}{>{\centering\arraybackslash}p{0.62cm}}

\definecolor{algcommentgreen}{RGB}{0,90,40}

\definecolor{unoorange}{RGB}{185,88,24}
\newcommand{\UnoDiff}[1]{{\color{unoorange}#1}}

\algrenewcommand{\algorithmiccomment}[1]{%
  \hfill{\color{algcommentgreen}\(\triangleright\)~#1}%
}

\title{Unlocking Lossless Speedups in LLMs \\ via Discrete Diffusion}

\newcommand{\temp}{\text{temp}}

\author{
Subham Sekhar Sahoo$^{*,\dagger,1}$, \quad Lingjie Chen$^{\dagger,1,2}$, \quad Khiem Pham$^{\dagger,1,3}$, \quad  Jonathan Geuter$^{\dagger,1,4}$, \quad  Chaitanya Dwivedi$^{1}$, \quad  Varad Pimpalkhute$^{1}$, \quad Yash Akhauri$^{1}$,\quad  Alexander Moreno$^{1}$, \quad Mikhail Yurochkin$^{1}$,\quad  Zhenting Wang$^{1}$,\quad  Mostafa Elhoushi$^{5}$, \quad Nolan Dey$^{5}$, \quad Shane Bergsma$^{5}$, \quad Joel Hestness$^{5}$, \quad John Thickstun$^{3}$, \quad Eric Xing$^{1}$, \quad Zhengzhong Liu$^{1}$
}

\affiliation{
    $^{1}$Institue of Foundation Models, $^{2}$University of Illinois Urbana-Champaign, $^3$Cornell Tech\\
    $^4$Harvard University $^5$Cerebras Systems \\
    $^\dagger$ Core Contributors \quad  $^*$ Correspondence to subham.sahoo@mbzuai.ac.ae 
}

\def\method{Uno}
\def\methodunosmall{Uno}

\def\methodqwen{Uno$_\text{\color{grayseq} Qwen}$}
\def\methodqwenc{{\color{zaibluefg} Uno}$_\text{{\color{grayseq} Qwen}}$}
\def\eagle{EAGLE-3}
\def\dflash{DFlash}
\def\wbase{{\theta_\text{AR}}}
\def\nld{\text{Nemotron-Labs-Diffusion}}
\def\dg{\text{DiffusionGemma}}
\def\wlora{{\theta_{\Delta}}}
\def\pdraft{p_\text{draft}}
\def\clean{\x^{L}}
\def\tps{\tau}
\def\methodqwenone{Uno$_\text{\color{grayseq} Qwen}^{\text{\color{grayseq}  1ep}}$}

\begin{document}

\begin{abstract}
Large Language Models (LLMs) owe much of their success to next-token prediction (NTP), but their autoregressive (AR) structure requires slow, sequential token generation. To overcome this bottleneck, we introduce diffusion-augmented LLMs, a new class of models that defines an AR model distribution while using diffusion to draw multiple tokens in parallel from that distribution. We decouple the parameters of these models into two sets: AR weights, trained using the standard NTP objective, and lightweight diffusion weights, trained to generate multiple tokens simultaneously. The diffusion weights are learned through a simple Diffusion Distillation phase that adds negligible overhead to existing LLM training pipelines. We also introduce $\Psi$-Spec, a family of samplers that enables lossless acceleration and inference-time scaling at a fixed context length.
Unlike speculative decoding, our method requires no separate draft model. Unlike diffusion LLMs (d-LLMs), it accelerates generation without sacrificing the quality of the underlying AR model. The resulting models, called Uno, can be trained from scratch or built by augmenting existing open-weight AR LLMs. Uno achieves higher throughput than leading speculative-decoding methods at every evaluated batch size and delivers up to $3\times$ speedups over the base AR model, including at the largest batch size supported by the device. Notably, our 8B Uno model outperforms the leading open d-LLM, the 26B DiffusionGemma, and the proprietary Mercury 2 across all evaluated benchmarks in agentic tool use, coding, and long-context reasoning.
We release code and checkpoints on:

\vspace{0.3ex}
\centerline{
\href{https://s-sahoo.github.io/uno/}{{https://s-sahoo.com/uno}}
}
\vspace{-1em}

\end{abstract}
\maketitle

\begin{figure}[H]
    \centering
    \vspace{-1.2em}
    \includegraphics[width=\linewidth]{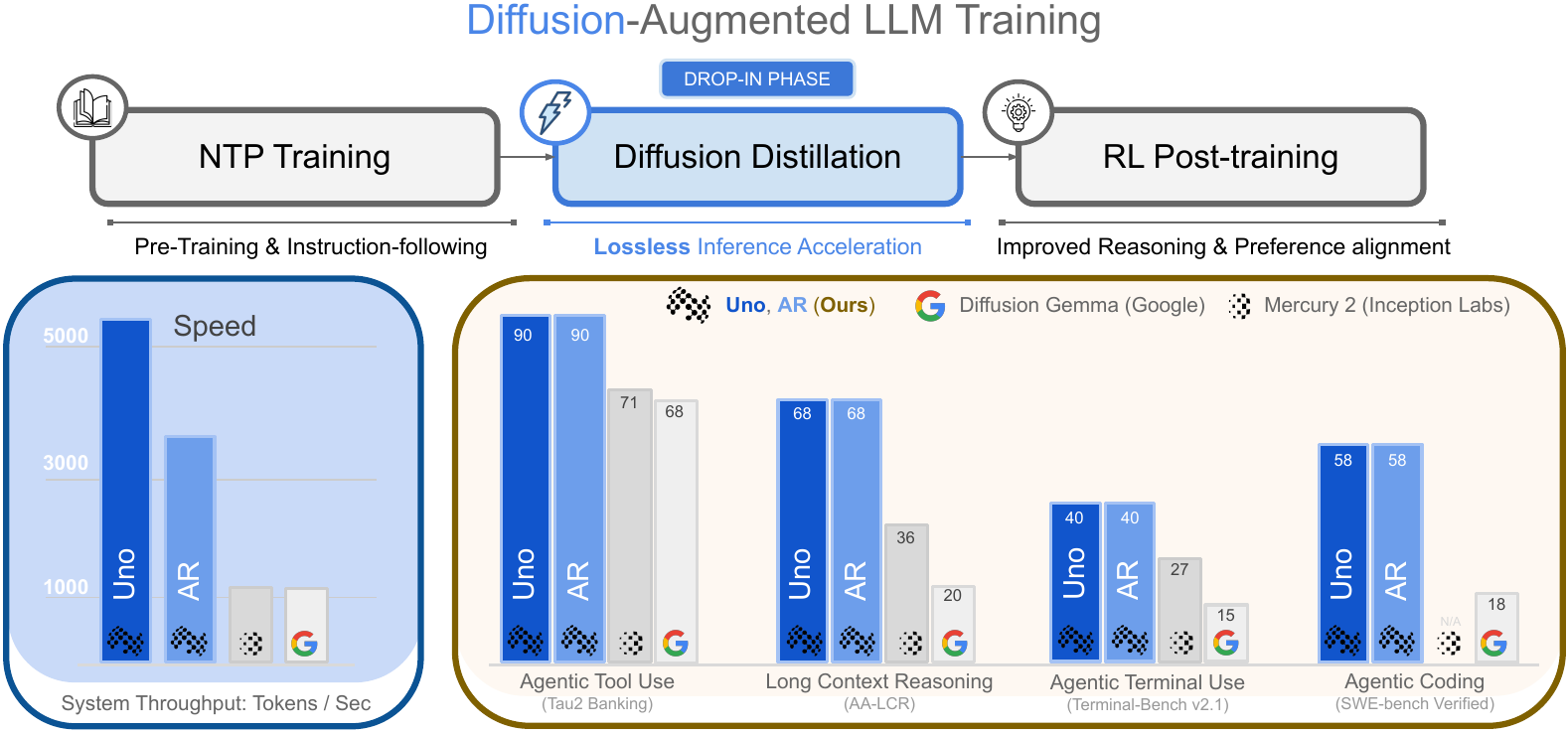}
    \caption{\textit{(Top)} Training overview for diffusion-augmented LLMs. Gray cells indicate AR-weight training, while the blue cell indicates diffusion-weight training. \textit{(Bottom Left)} System throughput of \method{}, the base AR model, and the baselines; see~\sec{sec:experiments:uno} for details. \textit{(Bottom Right)} Performance across agentic and long-context reasoning benchmarks.}
    \label{fig:opening-figure}
\end{figure}

\section{Introduction}\label{sec:introduction}
Large language models (LLMs) are increasingly showing human-level capabilities across coding, mathematics, and complex reasoning~\citep{team2026kimi,glm5team2026glm5vibecodingagentic}. These capabilities arise from a simple yet highly effective training objective: next-token prediction (NTP) over massive text corpora~\citep{brown2020language}. A key limitation of this objective is that inference is inefficient. By training models to predict only the next token, it produces autoregressive (AR) systems that generate just one token per decoding step. This sequential constraint becomes increasingly costly as reasoning traces grow longer, increasing serving latency and slowing reinforcement-learning (RL) post-training, where rollout generation dominates runtime~\citep{kim2026efficientrollout}.

Sequential decoding is poorly suited to both natural language and modern accelerators. Language contains predictable collocations and formulaic sequences that could be generated together in blocks~\citep{church1990word,biber2009corpus}, but standard LLMs cannot exploit this redundancy. Moreover, decoding is often \emph{memory bound}, especially at long context lengths, because moving model weights and key-value states limits inference speed and leaves GPUs underutilized~\citep{nvidia2023inference}. Predicting multiple tokens per step can amortize these memory transfers and better use parallel compute.

Existing approaches only partially address this opportunity. 
Speculative decoding accelerates generation by verifying tokens proposed by a smaller draft model~\citep{leviathan2023fast,chen2023accelerating}, but its gains depend on an efficient, well-aligned drafter and require maintaining a separate model. Discrete diffusion models~\citep{austin2021structured,sahoothesis} support parallel generation natively and have succeeded in biological domains~\citep{sahoo2024diffusion,schiff2025simple,tang2025peptune,lee2025genmol}, yet leading d-LLMs (d-LLMs) still face a quality-speed tradeoff relative to AR models~\citep{sahoo2025esoteric,sahoo2026scaling,bie2026llada21,diffusiongemma}. Their speedups also vanish at large inference batch sizes~\citep{wu2025fastdllmv2,fu2026nemotronlabsdiffusion,diffusiongemma}. 
The (lossy) speedups offered by d-LLMs at low batch sizes can therefore be insufficient in practice, as agentic workloads now predominate in contemporary LLM applications~\citep{google2025roi,aws2026agenticperformance}. In these workloads, a single request can trigger parallel agents, branching trajectories, tool calls, and retries, while serving systems batch calls across requests to improve accelerator utilization. \textbf{Batch-size-one latency therefore captures a narrow operating regime and may overstate speedups that diminish under concurrency}. Thus, practical acceleration should be evaluated at realistic batch sizes.

Our core idea is simple: define a high-quality AR distribution, then learn to sample multiple tokens in parallel from that same distribution. We realize this idea through 
\emph{diffusion-augmented LLMs} (\sec{sec:method}), which use a single architecture with two decoupled sets of weights: one governing response quality and the other optimized for generation speed.
Unlike multi-token prediction (MTP) approaches~\citep{cai2024medusa,gloeckle2024better}, which modify the LLM architecture by adding prediction heads for future tokens, our method augments each layer in the LLM with lightweight diffusion weights alongside the standard AR weights.  
Our training pipeline first trains the AR weights through NTP loss, then freezes them and learns the diffusion weights through \emph{Diffusion Distillation} (\sec{sec:method}) to generate token blocks in parallel with negligible training overhead. 
Our $\Psi$-Spec sampler (\sec{sec:samplers}) performs parallel prediction from the AR distribution. \textbf{The resulting model is a drop-in alternative to speculative decoding and self-speculative decoding}, requiring neither a separate draft model nor lossy AR-to-diffusion conversion.

We call our models \textbf{Uno} because they unify AR and diffusion weights within one architecture. Across our evaluations, Uno improves the speed-quality frontier over the speculative-decoding methods EAGLE-3~\citep{li2026eagle} and DFlash~\citep{chen2026dflash}, the open-weight d-LLMs Nemotron-Labs-Diffusion~\citep{fu2026nemotronlabsdiffusion} and DiffusionGemma~\citep{diffusiongemma}, and the proprietary d-LLM Mercury~2~\citep{ermon2026mercury2}. Its speedup over the base AR model persists at every evaluated batch size, including the largest batch that fits on the device with the AR weights, supporting both low-latency generation and high-throughput serving. Our contributions are:
\begin{enumerate}
    \item We introduce diffusion-augmented LLMs, which decouple the parameters that determine response quality from lightweight parameters optimized for generation speed, together with a corresponding training pipeline (\sec{sec:method}).
    \item We propose $\Psi$-Spec samplers to sample from diffusion-augmented LLMs. These samplers enable both lossless, AR-verified acceleration and inference-time scaling through additional denoising at a fixed context length; see \sec{sec:samplers}.
    \item We show how to train diffusion-augmented LLMs from scratch (\sec{sec:experiments:uno}) or create them by augmenting an existing open-weight AR LLM  (\sec{sec:experiments:qwen}).
    \item We show that \method{} achieves up to a $2\times$ speedup at the largest batch size supported by the base AR model, accelerating both inference and end-to-end RL training (\sec{sec:experiments:uno}).
\end{enumerate}

\section{Background}\label{sec:background}

\paragraph{Notation} We denote scalar discrete random variables with $K$ categories as `one-hot' column vectors and define $\onehotset = \{\mathbf{v} \in \{0, 1\}^K: \sum_{i=1}^K \mathbf{v}_i = 1\}$ as the set of all such vectors.
Define $\cat(\cdot\ ;\  \prior)$ as the categorical distribution over $K$ classes with probabilities given by $\prior \in \Delta$, where $\Delta$ denotes the $K-1$-simplex.
We also assume that 
the $K$-th category
corresponds to a special \texttt{[MASK]} token and let $\m \in \onehotset$ be the one-hot vector for this mask, i.e., $\m_{K} = 1$.
Additionally, let $\vone = \{1\}^{K}$, and
$\langle \mathbf{a}, \mathbf{b} \rangle$ and $\mathbf{a} \odot \mathbf{b}$ respectively denote the dot and Hadamard products between two vectors $\mathbf{a}$ and $\mathbf{b}$.
We use $\x \in \onehotset^L$ to represent clean data which is a length-$L$ sequence containing no mask tokens, and let $\xl$ denote its $\ell^{\text{th}}$ element where $\ell$ refers to the token index. Under our notation, each $\xl$ is a one-hot vector. 
The notation $\x^{m:n}$ denotes the subsequence from index $m$ to index $n$, inclusive, while $\x^{<\ell} \coloneqq \x^{1:\ell-1}$ denotes the prefix consisting of the first $\ell-1$ tokens. 
Finally, $[\cdot, \cdot]: \onehotset^m \times \onehotset^n \to \onehotset^{m+n}$ denotes the operator that concatenates sequences of lengths $m$ and $n$. Colors are used throughout the paper solely for emphasis and carry no mathematical meaning.

\subsection{Autoregressive Models}\label{sec:background:ar}

Consider a length-$L$ sequence $\x \in \onehotset^L \sim q_{\text{data}}$ drawn from the data distribution $q_{\text{data}}$.
Autoregressive (AR) language models factorize the joint distribution using the chain rule:
\begin{align}\label{eqn:ar-loss}
    \log p_\theta(\x)
    \;=\;
    \sum_{\ell=1}^{L} \log p_\theta(\x^\ell \mid \x^{<\ell}) .
\end{align}
Here, $p_\theta:\onehotset^{L} \to \Delta^{L}$ is typically implemented with a causal Transformer~\citep{vaswani2017attention} with parameters $\theta$.
This factorization leads to strong likelihood modeling and enables efficient inference primitives such as KV caching.
However, AR generation is intrinsically sequential: $\ell^\text{th}$ token can only be generated after the prefix $\x^{<\ell}$ has been generated.
As a result, producing an output of length $L$ requires $L$ causal decoding decisions.

\paragraph{Speculative Decoding} 
Speculative decoding \citep{leviathan2023fast,chen2023accelerating} uses a smaller draft model to autoregressively generate a block of candidate tokens, while a larger base model verifies them. The key observation is that although sampling from the base model is inherently sequential and expensive, evaluating the likelihood of an entire candidate sequence can be performed efficiently in parallel. Speculative decoding exploits this asymmetry by generating candidate tokens with the draft model that can be sampled more efficiently, and verifying them via rejection sampling in a single forward pass, accepting the longest valid prefix. This accelerates inference while exactly preserving the target  distribution.

\subsection{Discrete Diffusion Language Models}\label{sec:background:diffusion}

Discrete diffusion corrupts a clean sequence $\x \in \onehotset^L \sim q_{\mathrm{data}}$ into a simple prior $\prior^L$ and learns to reverse the corruption. We use the interpolating forward process from~\citet{sahoo2024simple}:
\begin{equation}\label{eqn:interpolating-forward}
    \z_t^{{\color{grayseq} \supl}}\sim q_t^{{\color{grayseq} \supl}}(\cdot\mid\x;{\prior})
    :=\cat\!\left(\cdot\ ;\  \at\x^{\color{grayseq} \supl}+(1-\at)\prior\right),
\end{equation}
where $\z_t$ denotes the noisy sequence at time step $t\in[0,1]$, and the noise schedule $\at$ decreases monotonically from $\alpha_0=1$ at clean data to $\alpha_1=0$ at the prior. A denoising model $\denoise:\onehotset^L\times[0,1]\rightarrow\Delta^L$, parameterized by $\theta$, predicts the clean tokens from $\z_t$ and $t$. Masked diffusion uses $\prior=\m$, while uniform-state diffusion uses $\prior=\vone/K$. We use the latter because it natively supports self-correction, few-step generation~\citep{sahoo2025the}, and better inference-time scaling than masked diffusion models~\citep{deschenaux2026the}.

\paragraph{Sampling}
For $0\leq s<t$, let $q_{s\mid t}$ denote the reverse posterior that maps $\z_t$ to a less noisy state $\z_s$. The $\Psi$-samplers proposed by~\citet{deschenaux2026the} form a family of discrete-diffusion samplers that generalize posterior (ancestral) sampling by incorporating predictor-corrector capabilities, thereby producing higher-quality samples. Their practical token-wise transition kernel is given by
{\small
\begin{align}
    [\Psi^\theta_{s\mid t}(\cdot\mid\z_t;\denoise(\z_t),{\color{grayseq}\prior})]^{\supl} =\kappa_t q_{s\mid t}^{\supl}(\cdot\mid\z_t,\denoise(\z_t);{\color{grayseq}\prior})
    +(1-\kappa_t)\left[
    \alpha_s q_{0\mid t}^{\supl}(\cdot\mid\z_t,\denoise(\z_t);{\color{grayseq}\prior})
    +(1-\alpha_s){\color{grayseq}\prior}
    \right],
    \label{eqn:psi-sampler}
\end{align}
}
where $\kappa_t\in[0,1]$ controls the strength of the correction. The exact functional form of \Eqn{eqn:psi-sampler} for MDMs and USDMs, see \supp{supp:background:psi-samplers}.

\paragraph{Discrete Consistency Distillation}
Discrete Consistency Distillation (DCD) compresses a multistep uniform-state diffusion process into a few-step generator~\citep{sahoo2025the}. 
It constructs a deterministic discrete trajectory from $\z_1 \sim \vone^L / K$ to $\x$ using the Gaussian diffusion process underlying the uniform-state discrete diffusion process,. 
For adjacent states $\z_t$ and $\z_s$, with $s<t$, the student distribution $\student^\ell(\z_t,t)$ is trained to match the teacher distribution $\teacher^\ell(\z_s,s); \forall \ell \in [L]$:
\begin{align}\label{eqn:dcd-loss}
    \mathcal{L}_{\mathrm{DCD}}(\studentw;\teacherw)
    =\sum_{\ell=1}^{L}
    \kl\!\left(\student^\ell(\z_t,t)\,\|\,\teacher^\ell(\z_s,s)\right).
\end{align}
Increasing the gap between $s$ and $t$ across distillation rounds teaches the student to take larger denoising steps. Refer \supp{supp:background:dcd} for further details.

\section{Diffusion-augmented LLMs}\label{sec:method}
We propose a new framework for designing LLMs that decouples generation quality from generation speed by augmenting each layer of a standard AR model with a separate set of diffusion weights dedicated to parallel token generation.
We call the resulting model a diffusion-augmented LLM. 
The key distinction is that each layer contains two sets of weights: AR weights, trained to generate tokens autoregressively and determine response quality, and diffusion weights, trained to generate tokens in parallel. This separation preserves the established AR training pipeline while introducing a dedicated phase for training the diffusion weights to accelerate inference. At generation time, both sets of weights draft tokens in parallel, after which the AR weights alone verify the drafts. Our sampler provably preserves the AR model's output distribution~(\sec{sec:samplers}), enabling lossless acceleration without sacrificing response quality. The framework can be applied to any causal autoregressive neural network, including causal Transformers~\citep{vaswani2017attention} and State-Space Models (SSMs; \citeauthor{gu2020hippo}, \citeyear{gu2020hippo}; \citeauthor{gu2023mamba}, \citeyear{gu2023mamba}). Because of this separation, the framework can also augment existing open-weight LLMs with parallel generation capabilities, providing lossless speedups without retraining their base parameters~(\sec{sec:experiments:qwen}).

The remainder of this section develops the framework in three stages. We first describe the roles of the AR and diffusion weights in a Diffusion-augmented LLM~(\sec{sec:method:architecture}). We then present the Diffusion Distillation Phase~(\sec{sec:method:diffusiondistillation}), which trains the diffusion weights using a one-step block-denoising formulation and associated training objectives. Finally, we discuss the training curriculum for two settings: accelerating inference alone and accelerating both RL rollouts and inference~(\sec{sec:method:practical-considerations}).

\subsection{Autoregressive and Diffusion Pathways}\label{sec:method:architecture}
Each layer of a diffusion-augmented LLM contains a set of autoregressive (AR) weights and a set of diffusion weights. These two sets of weights follow different training procedures, as described below.

\paragraph{Autoregressive Weights} We denote the base AR weights by $\wbase$. These parameters are trained using the standard LLM recipe of Pre-training~\citep{brown2020language,hoffmann2022training,grattafiori2024llama}, supervised fine-tuning (SFT;~\citeauthor{wei2021finetuned},~\citeyear{wei2021finetuned};~\citeauthor{chung2024scaling},~\citeyear{chung2024scaling}), and RL post-training~\citep{ouyang2022training, shao2024deepseekmath, guo2025deepseek}.
Given a length-$L$ sequence $\x \in \onehotset^L$, the model $\x_{\wbase}: \onehotset^L \to \Delta^L$ is causally masked, so its output at position $\ell$ observes only the prefix $\xll$, the distribution of the $\ell^\text{th}$ token is modeled as:
\begin{align}\label{eqn:denoise-ar}
    p_{\wbase}(\xl \mid \xll) = \x^{\ell - 1}_{\wbase}(\x), \qquad 1 < \ell \leq L.
\end{align}
We assume that the first token in $\x$ is a special $\texttt{<BOS>}$ token that requires no modeling.

\paragraph{Diffusion Weights} The diffusion weights $\wlora$ are used solely to accelerate generation. As described in \sec{sec:samplers}, our sampler uses the diffusion pathway to draft tokens and performs rejection sampling against the AR distribution defined by $\wbase$. We parameterize the diffusion weights as LoRA (Low RAnk) adapters~\citep{hu2022lora}: each AR weight matrix has an associated LoRA weight. Thus, the diffusion pathway uses $\wbase+\wlora$ to generate draft tokens, whereas the verification pathway uses $\wbase$, with the two distributions coupled through their shared base parameters. We train only $\wlora$ during the Diffusion Distillation Phase while keeping $\wbase$ frozen; see \sec{sec:method:diffusiondistillation}.

For effective rejection sampling, the draft distribution must remain closely coupled to the verification distribution. Parameterizing the draft pathway as a LoRA adaptation of the base model promotes this coupling while adding minimal inference-time memory overhead compared with a separate set of full-sized diffusion weights. This design retains the lossless verification of standard speculative decoding without requiring a separate draft model. At the same time, like self-speculative decoding, it tightly couples the drafting and verification pathways, but does so while remaining lossless..

Diffusion training teaches the LLM to refine an entire sequence in parallel. Specifically, given a noisy sequence $(\z_t \in \onehotset^L)_{t \in [0,1]}$ obtained by corrupting a clean sequence $\x \in \onehotset^L$ according to \Eqn{eqn:interpolating-forward}, the model is trained to recover $\x$. {Throughout this work, the LLM retains the NTP parameterization of standard autoregressive LLMs}: the logits at each position predict the subsequent clean token. This parameterization differs from that of conventional diffusion language models~\citep{austin2021structured,lou2024discrete,sahoo2024diffusion}, which typically predict the clean token at the same position. Accordingly, the distribution of the clean token $\xl$ is:
\begin{align}
    p_{\wbase,\wlora}(\xl \mid \z_t) = \x^{\ell - 1}_{\wbase,\wlora}(\z_t), \qquad 1 < \ell \leq L.
\end{align}

\subsection{Diffusion Distillation Phase}\label{sec:method:diffusiondistillation}
We train the diffusion parameters $\wlora$ to draft a block of tokens in parallel that the autoregressive (AR) model with parameters $\wbase$ would generate sequentially. Specifically, our goal is to match the single-step diffusion distribution $p_{\wbase,\wlora}(\cdot)$ over a token block to the AR distribution $p_{\wbase}(\cdot)$ over the same block. Discrete Consistency Distillation (DCD;~\sec{sec:background:diffusion}) is well suited to this task because it can distill a multistep diffusion process into a few-step generator while closely approximating the distribution induced by the original process.

In what follows, we first describe how we adapt DCD to approximate the AR distribution through single-step diffusion. We then extend the DCD training objective, namely the single-step distillation loss $\mathcal{L}_{\text{DCD}}$, with another auxiliary loss term: $\mathcal{L}_{\text{TV}}$ to encourage longer diffusion drafts to be accepted by the AR verifier. The resulting objective is:
\begin{align}\label{eqn:training-loss}
    & \gL(\wlora; \wbase, \alpha, \beta) = \E_{\x \sim \gD, \z_1 \sim \prior{}^L} \left[
        \alpha {\mathcal{L}_{\text{DCD}}(\wlora; \wbase, \x, \z_1)}
        + \beta {\mathcal{L}_{\text{TV}}(\wlora; \wbase, \x, \z_1)}\right],
\end{align}
where $\z_1$ is the fully corrupt sequence.
Although $\alpha=0,\beta=1$ yields the largest speedups, training first with $\alpha=\beta=1$ and then switching to $\alpha=0,\beta=1$ accelerates convergence. We ablate both loss terms in \sec{sec:experiments:ablation}. We describe each term below.

\paragraph{One-step Distillation} 
Standard DCD constructs a multistep denoising trajectory by simulating intermediate states of the underlying PF-ODE. At LLM scale, repeatedly constructing and storing these intermediate Gaussian latents is prohibitively expensive. We therefore distill the entire trajectory into a single denoising step that maps a fully corrupted input $\z_1 \sim \prior{}^L$ directly to the clean sequence $\x$. The frozen base model defines the autoregressive teacher distribution $\x_{\wbase}$, and the adapted model defines the student distribution $\x_{\wbase,\wlora}$. We train the student to match the teacher directly, eliminating the need to simulate or materialize any intermediate PF-ODE states.

\paragraph{Block Diffusion} Recovering an entire long sequence from a fully corrupted input in a single denoising step is prohibitively difficult; we therefore perform one-step denoising blockwise. We partition the clean sequence $\x$ and the fully corrupt sequence $\z_1$ into $N$ blocks of size $B$. We use $\x^{(b)}$ and $\z_1^{(b)}$ to denote the $b^\text{th}$ blocks of $\x$ and $\z_1$, respectively.
For each block, we train $\wlora$ to match the student diffusion distribution over the clean block $\x^{(b)}$, conditioned on the noisy block $\z_1^{(b)}$ and the preceding clean context $\x^{(<b)}$, to the teacher AR distribution over the same block, conditioned on the preceding clean context. We compute the teacher and student predictions in a single forward pass over the concatenated sequence $[\x,\z_1]$. Specifically, we use a block-causal attention mask that permits causal attention within $\x$ and within each noisy block $\z_1^{(b)}$. Tokens in $\z_1^{(b)}$ additionally attend to all preceding clean blocks $\x^{(<b)}$.

The main challenge is to compute the teacher logits at positions in $\x$ using only $\wbase$, while computing the student logits at positions in $\z_1$ using both $\wbase$ and $\wlora$. We achieve this using gated LoRA~\citep{samragh2025your}, which disables the adapters at clean-sequence positions and enables them at noisy-sequence positions. Consequently, $\wbase$ produces the teacher logits, whereas $\wbase$ and $\wlora$ jointly produce the student logits.
The resulting blockwise DCD objective is:
\begin{align}\label{eqn:dcd-block}
    \mathcal{L}_{\text{DCD}}(\wlora; \wbase, \x, \z_1)  =  \sum_{b = 1}^N \sum_{\ell = 1}^{B} 
        \kl \left(\x^{(N +  b, \ell)}_{{\wlora,\wbase}} ([\x, \z_1]) \| \x^{(b, \ell)}_{{\wbase}} ([\x, \z_1]) \right).
\end{align}
The base parameters $\wbase$ remain frozen throughout fine-tuning, and only the LoRA parameters $\wlora$ are updated.

\paragraph{Total Variation Loss} 
Our sampler~(\sec{sec:samplers}) performs rejection sampling against the AR distribution and retains only the longest consecutive prefix of the diffusion prediction. Sampling efficiency therefore depends on the length of this retained prefix. To increase its expected length, we minimize the blockwise Total Variation (TV) distance between the diffusion and the AR distributions following Corollary~3.6 of~\citet{leviathan2023fast}:
\begin{align}\label{eqn:tv-loss}
    {\mathcal{L}_{\text{TV}}(\wlora; \wbase, \x, \z_1)} =  \sum_{b = 1}^N \sum_{\ell = 1}^{B}  \left|\x^{(N + b, \ell)}_{{\wlora,\wbase}} ([\x, \z_1]) - \x^{(b, \ell)}_{{\wbase}} ([\x, \z_1]) \right|.
\end{align}
Minimizing this objective increases the probability of consecutive tokens being accepted, and, consequently, increases the expected length of the accepted draft prefix.

\subsection{Practical Considerations}\label{sec:method:practical-considerations}
The AR weights $\wbase$ and diffusion adapters $\wlora$ serve complementary roles and are optimized using separate objectives, making their training order an important practical consideration. In particular, Diffusion Distillation can be performed either after all AR training is complete or before RL post-training so that the resulting adapters can accelerate rollout generation. We therefore consider two training regimes, depending on whether diffusion-based generation is used only during inference or during both RL post-training and inference.

\paragraph{Faster Inference Only} 
If the sole objective is faster inference, we first complete autoregressive pre-training and post-training, or simply initialize them with an Open-weights LLM. We then freeze the AR weights $\wbase$ and train $\wlora$ using Diffusion Distillation (\sec{sec:method:diffusiondistillation}).

\paragraph{Faster RL Training \& Faster Inference} 
We demonstrate in \sec{sec:experiments} that our method provides significant generation speedups wrt the base AR model across all batch sizes, and can therefore accelerate RL post-training due to faster rollouts. In this setting, we first apply diffusion distillation after supervised fine-tuning and before RL, as illustrated in~\Fig{fig:opening-figure} (top). The resulting adapters can then accelerate rollouts during RL.
Standard RL policy-optimization recipes, including PPO and GRPO-based methods~\citep{ouyang2022training,shao2024deepseekmath,guo2025deepseek,yu2026dapo}, update $\wbase$ but not the $\wlora$. 
One might therefore expect that, as $\wbase$ changes, the draft distribution will drift away from the verifier distribution, reducing the acceptance rate and eroding the speedup. 
\textbf{Surprisingly, we show in \sec{sec:experiments:uno} that the speedup is retained even as the AR weights are trained using RL.}

\section{\texorpdfstring{$\Psi$}{Psi}-{Speculative} Sampler}\label{sec:samplers}
Our goal is to sample from the AR distribution defined by $\wbase$ while drawing multiple tokens in parallel. We therefore introduce the $\Psi$-Speculative sampler ($\Psi$-Spec), which uses the diffusion pathway to propose a block of tokens in parallel~(\sec{sec:samplers:draft}) and performs rejection sampling against the base AR distribution to accept the longest valid prefix~(\sec{sec:psi:verify}).

\subsection{Diffusion Sampler}\label{sec:samplers:draft} 
Let $\x \in \onehotset^{L}$ denote a partially generated sequence of length $L$. To generate a block of $B$ tokens, we first append $B-1$ random tokens sampled from the prior, $\z_1 \sim \prior^{B-1}$. Denoising then proceeds from $t=1$ to $t=0$, with $\z_t$ denoting the partially denoised sequence at time $t$. At each step, the denoiser
$\x_{\theta}(\cdot\ ;\   \x^{<L}):\onehotset^{B} \to \Delta^{B}$
operates on the block $[\clean, \z_t] \in \onehotset^B$, while the activations for the preceding sequence $\x^{<L}$ remain in the KV cache. In \sec{sec:samplers:draft:proposal}, we define the proposal distribution over a block of tokens, and in~\sec{sec:samplers:draft:sample}, we describe how candidates are sampled from this distribution.

\subsubsection{Diffusion Proposal Distribution}\label{sec:samplers:draft:proposal}
Given a noisy block $\z_t$, we use the $\Psi$-Spec transition in \Eqn{eqn:psi-sampler} to sample a less noisy block $\z_s$, where $s<t$. Because the denoiser uses the NTP parameterization, the clean-token distributions used to construct $\z_s$ are given by $\x^{1:B-1}_{\wbase, \wlora}([\clean, \z_t])$. 
The input contains both clean and corrupted tokens, while the diffusion weights are trained only on corrupted tokens. To avoid distribution shift, we compute the logits for the first, clean position using only the base AR weights, and those for the noisy positions using both the AR and diffusion weights. We achieve this in a single forward pass using the gated LoRA technique of \citet{samragh2025your}.

Let $\Psi_0$ denote the distribution induced by the $\Psi$-sampler at $t=0$. At the final step, we sample the first token using only the base AR weights:
\begin{align}\label{eqn:psispec-draft-z0} 
    \z_0^1 \sim \Psi^{\ell=1}_{0} \left(\cdot \mid \z_t ; \x^{{\color{grayseq} 1:B-1}}_{\wbase}(\cdot),   {\color{grayseq} \prior{}} \right).
\end{align}
We sample the remaining $B-2$ tokens from the following joint distribution, discussed in~\sec{sec:samplers:draft:sample}:
\begin{align}\label{eqn:psispec-draft}
    \z_0^{2:B} \sim \prod_{\ell = 2}^B \Psi^{\ell}_{0} \left(\cdot \mid \z_t ; \x^{{\color{grayseq} 1:B-1}}_{\wbase, \wlora}(\cdot),   {\color{grayseq} \prior{}} \right).
\end{align}

\subsubsection{Sampling Candidates}\label{sec:samplers:draft:sample}
Given the proposal distribution in~\Eqn{eqn:psispec-draft}, we construct a candidate set $\mathcal{C}=\{\c:\c\sim \prod_{\ell = 2}^B \Psi^{\ell}_{0} \}$. The number of draft candidates determine the tradeoff between acceptance length and verification cost. Sampling more candidates increases the probability of accepting a longer prefix, but also increases the cost of verification. We therefore select the largest candidate set that can be verified without reducing inference throughput, subject to the available serving batch size.

\paragraph{Linear Sampler (System Throughput Optimized)}
The simplest approach samples each draft token directly from its marginal distribution in~\Eqn{eqn:psispec-draft}, producing a single candidate sequence. At high batch sizes, inference can become compute-bound, leaving little spare compute for verifying additional candidates. The linear sampler is therefore well suited to maximizing aggregate system throughput in this regime.

\paragraph{Tree Sampler (Single-user Throughput Optimized)} 
At low batch sizes, inference is typically memory-bound, leaving substantial compute capacity underutilized. We exploit this spare compute by sampling multiple candidates and verifying them in parallel. Specifically, we use the tree-based sampling procedure of~\citet{cai2024medusa}, which selects the top $K$ tokens at each position within a block. Rather than evaluating all $K^{B-1}$ candidate sequences, we adopt the candidate-pruning strategy of~\citet{li2026eagle} which ranks candidates by their log-probabilities and retains the top $V\in\mathbb{Z}^{+}$ prefixes. Thus, $(B,K,V)$ are the hyperparameters of the tree sampler, corresponding to the block size, branching factor, and prefix budget, respectively.

\subsection{One-step Diffusion w/ AR Verification}\label{sec:psi:verify}
The central goal of this work is to accelerate generation. We therefore draft an entire block of $B$ tokens using a single diffusion forward pass. In~\supp{supp:subsec:singlestep}, we show that, for single-step diffusion generation, Eqns.~(\ref{eqn:psispec-draft-z0}, \ref{eqn:psispec-draft}) reduce to
\begin{align}
    & \z_0^1 \sim \x^{1}_{\wbase}(\cdot), \label{eqn:psispec-draft-z0-onestep}  \\
    & \z_0^{2:B} \sim \pdraft = \prod_{\ell = 2}^B \x^{\ell}_{\wbase, \wlora}(\cdot). \label{eqn:psispec-draft-onestep}
\end{align}
From $\pdraft$, we sample candidates $\mathcal{C}=\{\c:\c\sim\pdraft\}$ and apply the standard speculative-decoding rejection correction~\citep{leviathan2023fast}, retaining the longest prefix accepted by the base AR model. This preserves the base model's target distribution. For $|\mathcal{C}|>1$, candidates are verified concurrently as a prefix tree using tree attention~\citep{cai2024medusa}. Because $\wbase$ remains frozen and only the diffusion LoRA adapters are trained, the base AR model provides an unchanged verifier and enables lossless speculative speedups. The complete sampling algorithm is provided in~\alg{alg:psi-spec}.

\paragraph{Tokens-Per-Forward-pass (TPF)}
During drafting, the first token is generated using the base AR weights and therefore matches the verifier distribution exactly, so it is always accepted. If a subsequent draft token is rejected, the verifier samples a replacement from the renormalized residual distribution, as in standard speculative decoding. Consequently, even an immediate rejection produces two output tokens: the always-accepted first token and the verifier-sampled replacement token. If all $B$ draft tokens are accepted, the verifier additionally samples one token from the logits following the final draft token, producing $B+1$ output tokens. Because each iteration requires two forward passes, one for drafting and one for verification, the TPF is bounded by $1 \leq \mathrm{TPF} \leq \frac{B+1}{2}$.

We provide support for both the Linear and Tree samplers in Nano-vLLM~\citep{yu2025nanovllm} and SGLang~\citep{zheng2024sglang}. All experiments reported in this paper were conducted using our Nano-vLLM implementation.

\subsection{Inference-Time Scaling}
$\Psi$-Spec introduces an additional axis for inference-time scaling in LLMs by allowing the number of denoising steps $T$ to exceed the number of drafted tokens $B$. \citet{deschenaux2026the} show that, for $0 \leq \kappa_t < 1$, increasing $T$ consistently improves sample quality. The central question is whether this improvement can eventually surpass the quality of AR generation. If so, AR verification should be disabled, since it would constrain the final output to the lower quality level of the AR model and prevent the gains from additional denoising from being realized. \textbf{Unlike conventional inference-time scaling methods, this approach allocates additional computation without increasing the context length.} 
We leave a systematic exploration of this quality-compute tradeoff to future work.

\section{Experiments}\label{sec:experiments}
We train diffusion-augmented LLMs in two settings. In the first, we train the AR weights from scratch on proprietary data and train the diffusion weights on data drawn from the same distribution. In the second, we augment an open-weight AR model, Qwen3-8B~\citep{yang2025qwen3}, without access to its original training data. We instead train the diffusion weights on a different data distribution using the open-source OpenThoughts dataset~\citep{guha2025openthoughts}. Together, these experiments demonstrate the versatility of our method and show that the diffusion and AR weights need not be trained on the same data distribution.

\paragraph{Throughput Analysis} 
Throughput depends strongly on context length. However, existing evaluations~\citep{wu2025fastdllmv2, fu2026nemotronlabsdiffusion} measure throughput on downstream tasks whose reasoning traces vary in length. This can misleadingly favor models that generate shorter traces. Following standard LLM evaluation practice\footnote{\url{https://nvidia.github.io/TensorRT-LLM/developer-guide/perf-benchmarking.html}}, we instead evaluate every method using $m$ random input tokens and a fixed output length of $n$ tokens. For diffusion and speculative decoding methods, which draft a block of $B$ tokens per forward pass, we first measure the average number of accepted tokens per forward pass (TPF) across all benchmarks. We then run $\left \lceil \frac{n}{\mathrm{TPF}}\right \rceil $ decoding steps, constraining each method to accept $\mathrm{TPF}$ tokens per step on average. This ensures that all methods are evaluated at the same input and effective output lengths. Throughout the paper, we use $m=1024$ and $n=8192$ and refer to this setting as the ``1K/8K throughput test''. 

\paragraph{Per-request Throughput vs. System Throughput}
We report throughput at batch size 1 and at the largest batch size that fits on a single H200 GPU. \textbf{Batch-size-1 throughput is less representative of practical serving because agentic workloads often generate concurrent requests even while serving a single user} as described in~\sec{sec:introduction}. We therefore also report system throughput, which is the aggregate token generation rate at the maximum feasible batch size, which better reflects serving capacity and cost efficiency. For each regime, we select the TPF setting that maximizes throughput and use the 1K/8K test described above.

\subsection{End to End Diffusion Augmented LLM Training}\label{sec:experiments:uno}

\subsubsection{Setup}
\paragraph{Benchmarks}
We evaluate on both agentic and non-agentic benchmarks. The agentic suite comprises the Telecom, Airline, and Retail domains of $\tau^3$-Bench~\citep{shi2026tau}, $\tau^2$-Bench~\citep{barres2025tau2} , Terminal-Bench v2.1~\citep{terminalbench2026v21}, and SWE-bench Verified~\citep{jimenez2024swebench}. The non-agentic suite includes AA-Omniscience~\citep{jackson2025aaomniscience}, AA-LCR~\citep{artificialanalysis2025aalcr}, Humanity's Last Exam (HLE;~\citeauthor{phan2025humanity},~\citeyear{phan2025humanity}) , GPQA-Diamond~\citep{rein2023gpqa}, GSM8K~\citep{cobbe2021gsm8k}, MATH500~\citep{lightman2024verify}, AIME 2024 and 2025~\citep{balunovic2025matharena}, AIME 2026~\citep{dekoninck2026matharena}, and MBPP~\citep{austin2021mbpp}. We use a context window of $262,144$ with $131,072$ maximum generation tokens, allowing for long prompts on agentic benchmarks, and sampling parameters $\temp=1$, top-$p=0.95$, and top-$k=50$.

\paragraph{Metrics} 
For each method, we report average pass@1 over multiple generations and average TPF; see~\tab{tab:benchmark-overview} for details. We compute throughput using the ``1k/8k throughput test,'' averaging TPF across benchmarks. System throughput is measured using the largest batch size that fits on a single H200 GPU, whereas per-request throughput is measured at batch size~1. 

\paragraph{Baselines}
We compare against \nld{}-14B~\citep{fu2026nemotronlabsdiffusion}, \dg{}-26B-A4B~\citep{diffusiongemma}, and Mercury~2~\citep{ermon2026mercury2}. \nld{} is pretrained autoregressively on 1T tokens and then fine-tuned as a diffusion model on 300B tokens. It also uses masked diffusion with bidirectional attention within draft blocks. We evaluate the largest \nld{} variant because it performs better on benchmarks than the smaller variants. \dg{} is a sparse 26B mixture-of-experts model with 4B active parameters, fine-tuned from Gemma~4~\citep{diffusiongemma}. Since \textbf{both \nld{} and \dg{} modify the base AR parameters, they are lossy unlike \method{}}. We use the default baseline configurations: \nld{} uses Linear Self-Speculation with a block size of 32, a maximum generation length of 8192, a thinking-token budget of 6000, and temperatures of 0 for drafting and verification. \dg{} uses a block size of 256, temperature 1, a maximum generation length of 131072, a context length of 162144, and an entropy-bounded diffusion sampler with a bound of 0.1 and thinking enabled.  We also compare against Inception Labs' closed-source d-LLM, Mercury~2, whose parameter count is undisclosed.

\subsubsection{\method{}: Diffusion Augmented LLM (Ours)}
We first define the architecture of our model. The weights corresponding to every layer are the the AR weights $\wbase$. Corresponding to every weight matrix, we introduce rank-128 LoRA adapters with LoRA-$\alpha=256$ which serve as the diffusion weights $\wlora$. We share more details subsequently.

\paragraph{LLM Architecture}
We use a dense decoder-only causal Transformer with 36 layers, hidden width
$4,096$, SwiGLU-style MLP width $12,288$, $32$ query heads, $8$ KV heads, head dimension
$128$, grouped RMSNorm, RoPE ($\theta=10^7$), a $250,624$-token vocabulary, and a maximum
configured context length of $524,288$ tokens. The model contains $6.95$B transformer-body parameters
plus approximately $2.05$B parameters from the large untied input/output vocabulary
matrices.

\paragraph{Training AR weights}
The AR weights are trained on approximately 23T tokens on internal quality data with a staged context-length extension schedule. Pretraining uses 21.9T tokens at an 8K-token context length, with a peak learning rate of $3 \times 10^{-4}$ under a warmup-stable-decay (WSD) configuration and a 3,000-step warmup. We then extend the context length to 32K, training on an additional 1.1T tokens while linearly decaying the learning rate from $3 \times 10^{-4}$ to $3 \times 10^{-5}$, following a 1,250-step warmup. Next, the context length is increased to 128K and the model is trained on 0.5T tokens at a constant learning rate of $3 \times 10^{-5}$, with a 500-step warmup. Finally, the context length is extended to 512K and trained for 0.3T tokens with a constant learning rate of $3 \times 10^{-5}$ and a 200-step warmup.

\paragraph{Training Diffusion Weights}
Diffusion weights are implemented as rank-128 LoRA adapters with LoRA-$\alpha=256$. We train them on 7B tokens sampled randomly from the SFT training data while keeping the AR weights frozen. 
We employ curricula for context length and diffusion block size: 1.8B tokens at context length $16,384$, divided into increasing block sizes of two, four, and eight with 600M tokens each; followed by 5.2B tokens at context length $65,536$ and block size 8. We train on a global batch size of $128$ with a WSD learning rate scheduler with 200 warmup steps and a peak learning rate of $5 \times 10^{-5}$.
In~\Eqn{eqn:training-loss}, we set $\alpha=0.01, \beta=1$. 
Training takes approximately 60 hours on 8 H200 nodes with 8 GPUs each.

To sample from \method{}, we use the Linear Sampler with $B=4$ to maximize system throughput and the Tree Sampler with $(B,K,V)=(16,32,32)$ to maximize single-request throughput at batch size 1.

\subsubsection{Results}

\paragraph{$\Psi$-Spec Sampler Configurations}
We examine how TPF and throughput vary across Linear and Tree sampler configurations in \tab{tab:uno-sampler-ablations}. For the Linear sampler, TPF increases from $1.8$ at $B=4$ to $2.2$ at $B=8$, but reaches only $2.3$ at $B=16$; $B=4$ achieves the highest system throughput. For the Tree sampler, with $B=16$ and $K=32$, $V=32$ provides higher per-request throughput than $V=64$, likely because the larger verification cost reduces throughput.

\paragraph{\method{} vs. Base AR} 
\method{} qualitatively matches the base AR model while achieving higher throughput at every batch size. In the 1k/8k throughput test, the largest batch size supported by the base AR model is 64, at which \method{} is $1.5\times$ faster (see \tab{tab:uno-baseline-throughput}). This speedup benefits multi-user serving and RL post-training. At batch size 1, \method{} is approximately $2.2\times$ faster.

\begin{remarkbox}
\textbf{Takeaway 1:} \method{} matches the qualitative performance of Base AR while delivering over $2\times$ higher throughput across all batch sizes.
\end{remarkbox}

\paragraph{Comparison with Open Weights Diffusion Language Models}
As shown in \tab{tab:eval-results}, \method{} outperforms the open-weight \dg{} and \nld{} models on all tasks, with larger gains on agentic tasks. Using the TPF values in \tab{tab:eval-results}, we measure maximum system throughput at the largest feasible batch size and per-request throughput at batch size 1. Despite using full attention in every layer, \method{} achieves higher system throughput (with Tree Sampler; $(B,K,V)=(16,32,32)$) than both baselines. Notably, \dg{} uses strided attention, with five local sliding-window self-attention layers for every global self-attention layer, yet remains slower at large batch sizes. \dg{} is faster at batch size 1 but has substantially lower accuracy, while \nld{} trails \method{} in both throughput and output quality.
Although smaller blocks may improve throughput, further tuning offers no practical benefit given the baselines' lower quality. Also, \textbf{note that both baselines are slower than their respective base AR models at large batch sizes}~\citep{fu2026nemotronlabsdiffusion, diffusiongemma}. 

\begin{remarkbox}
\textbf{Takeaway 2:} \method{} outperforms open-weight d-LLMs on every benchmark, while also achieving the highest system throughput.
\end{remarkbox}
\paragraph{Comparison with Proprietary Diffusion Language Models} 
As shown in \tab{tab:eval-results}, the 8B-parameter \method{} outperforms Mercury~2 on all Agentic Tool Use, Agentic Coding, and Long-Context Reasoning benchmarks, trailing only on one Science and Knowledge benchmark, potentially due to differences in model size and training data. More importantly, Mercury~2\footnote{{\tiny \url{https://artificialanalysis.ai/models/mercury-2?speed=output-speed-by-prompt-type\#speed-tabs}}} reports a system throughput of 1,154 tokens/s for a 1K-token input at batch size 10. \method{} achieves a maximum system throughput $\sim4.6\times$ higher, despite Mercury~2 running on substantially faster Blackwell GPUs. Moreover, Mercury~2 does not disclose its quantization, so it may also benefit from lower precision, whereas \method{} and the other baselines use bfloat16. This result highlights the strength of diffusion-augmented LLMs.
\begin{remarkbox}
\textbf{Takeaway 3:} \method{} beats proprietary d-LLM Mercury 2, on all agentic, coding, and long-context benchmarks and achieves $\sim4.6\times$ higher maximum system throughput, despite running on slower hardware.
\end{remarkbox}   

\paragraph{Faster RL Training} 
From the supervised fine-tuning (SFT) checkpoint, we trained four experts in mathematics, code generation, tool use, and web search using the DAPO reinforcement learning algorithm. During this stage, we updated only the base AR weights; the diffusion weights trained on the SFT checkpoint remained frozen and were used only to speed up RL rollouts. This yielded up to a $40\%$ end-to-end training speedup, primarily for the mathematics and code experts. Gains were smaller for the tool-use and search experts because tool calls dominated their runtime. We will provide detailed results in the next revision.
We then consolidated the four experts into a single model using ISO-Merger (RAM; \citeauthor{yuan2026behavior}, \citeyear{yuan2026behavior}), a data-free method that combines specialists trained from a shared base checkpoint without additional rollouts. As shown in \tab{tab:sft-rl-tpf-results}, the diffusion adapters trained on the SFT checkpoint retain their speedups after RL post-training, with only a nominal 6\% decrease in TPFs.

\begin{table*}[t]
\centering
\small
\setlength{\tabcolsep}{3pt}
\caption{
Accuracy and TPF of \method{}, \nld{}, Mercury 2, and \dg{} on agentic and non-agentic benchmarks. Mercury 2 results are from Artificial Analysis. Results for \nld{} and \dg{} are computed from their open-source checkpoints (\tab{tab:checkpoints}). For \method{}, subscripts report ``TPF$_1$ / TPF$_2$,'' where TPF$_1$ uses the system-throughput-optimal Linear sampler with $B=4$, and TPF$_2$ uses $(B,K,V)=(16,32,32)$, optimized for per-request throughput. $^*$Reported by Artificial Analysis's live tracker on August 30, 2026. 
}
\label{tab:eval-results}
\begin{tabular}{lZccc}
\toprule
& \normalZ{{\color{zaibluefg} \methodunosmall{}}}
& {Mercury 2}
& {{Diffusion}-{Gemma}}
& {{Nemotron-Labs}-{Diffusion}} \\
\textit{Model size}
& \normalZ{$(8\mathrm{B})$}
& N/A
& $(26\mathrm{B}\text{-}\mathrm{A}4\mathrm{B})$
& $(14\mathrm{B})$ \\
\midrule
\textit{Agentic Tool Use} & & & & \\
\quad $\tau^3$ Banking
& $\mathbf{25.8}_{1.8/2.7}$ & $9_{\na}$ & -- & -- \\
\quad $\tau^2$ Telecom
& $\mathbf{90.1}_{1.7/2.1}$ & $71_{\na}$ & $68.1_{18.8}$ & $14.3_{4.8}$ \\
\quad $\tau^2$ Retail
& $\mathbf{67.1}_{1.8/2.4}$ & -- & $65.5_{23.7}$ & $5.6_{3.1}$ \\
\quad Terminal-Bench v2.1
& $\textbf{39.6}_{2.1/2.7}$ & $27_{\na}$ & $14.7_{14.1}$ & $4.5_{7.5}$ \\
\textit{Agentic Coding} & & & & \\
\quad SWE-bench Verified
& $\mathbf{68.4}_{2.2/3.1}$ & -- & $18.7_{5.6}$ & $0.8_{1.5}$ \\
\textit{Long-Context Reasoning} & & & & \\
\quad AA-LCR
& $\mathbf{68.0}_{1.8/2.6}$ & $36_{\na}$ & $19.7_{10.8}$ & $7.3_{1.1}$ \\
\textit{Science and Knowledge} & & & & \\
\quad Humanity's Last Exam
& $\mathbf{18.6}_{1.8/2.8}$ & ${16}_{\na}$ & $9.2_{14.1}$ & $2.6_{7.2}$ \\
\quad GPQA-Diamond
& $\mathbf{77.1}_{2.0/4.2}$ & $77_{\na}$ & $70.7_{11.9}$ & $40.4_{7.6}$ \\
\quad AA-Omniscience
& $14.3_{1.7/3.1}$ & $\mathbf{20}_{\na}$ & $17.7_{9.9}$ & $11.0_{11.1}$ \\
\textit{Math} & & & & \\
\quad GSM8K
& $\mathbf{95.4}_{1.9/2.7}$ & -- & $95.1_{28.9}$ & $93.1_{6.1}$ \\
\quad MATH500
& $\mathbf{98.9}_{1.9/2.4}$ & -- & $92.4_{24.1}$ & $89.2_{5.6}$ \\
\quad AIME-24
& $\mathbf{93.0}_{1.9/2.5}$ & -- & $73.7_{16.9}$ & $56.7_{4.9}$ \\
\quad AIME-25
& $\mathbf{90.7}_{1.8/2.4}$ & -- & $74.3_{18.9}$ & $40.0_{4.5}$ \\
\quad AIME-26
& $\mathbf{86.3}_{1.8/2.4}$ & -- & $70.7_{17.8}$ & $46.7_{4.8}$ \\
\textit{Coding} & & & & \\
\quad MBPP
& $\mathbf{84.1}_{1.8/2.4}$ & -- & $80.1_{15.5}$ & $73.8_{5.3}$ \\
\quad HumanEval
& $\mathbf{95.2}_{1.9/3.4}$ & -- & $95.1_{28.2}$ & $84.8_{7.5}$ \\
\midrule
Avg. TPF & 1.9 / 2.7 & \na{} & 17.56 & 5.41 \\
System Throughput & \textbf{5255} & $1197^*$ & 1136 & 2794 \\
Per-request Throughput & 405 & $769^*$ & \textbf{836} & 290 \\
\bottomrule
\end{tabular}
\end{table*}
\subsection{Open Weights LLMs}\label{sec:experiments:qwen}
We show that Diffusion-augmented LLMs can initialize their autoregressive (AR) weights from the open-weight Qwen3-8B model while retaining the speedups enabled by diffusion weights trained on a different data distribution, specifically, the open-source OpenThoughts dataset. Although training Qwen3-8B on this dataset degrades its quality (see~\tab{tab:qwen-openthoughs-ablation}), the diffusion weights trained on the same nevertheless enable lossless speedups on the same benchmarks.

\subsubsection{Setup}

\paragraph{Benchmarks}
We evaluate on mathematical reasoning (GSM8K, MATH500, AIME-24, AIME-25, and AIME-26), code generation: HumanEval~\citep{chen2021humaneval}, MBPP~\citep{austin2021mbpp}, and LiveCodeBench v6~\citep{jain2025livecodebench}); Science reasoning: GPQA and GPQA-Diamond~\citep{rein2023gpqa}; instruction following: IFEval with prompt-level strict accuracy~\citep{zhou2023ifeval}, and General Knowledge: MMLU-Pro~\citep{wang2024mmlupro}. 

\paragraph{Lossless Baselines (Speculative Decoding Methods)}
Our primary lossless baselines are EAGLE-3~\citep{li2026eagle}, which uses an AR drafter, and \dflash{}~\citep{chen2026dflash}, which uses a diffusion drafter. Both draft models are designed for Qwen3, and we evaluate their open-source checkpoints.
EAGLE-3 adds a lightweight $0.40\mathrm{B}$-parameter drafter that sequentially predicts tokens from fused intermediate features of the target model. \dflash{} instead adds a $1.05\mathrm{B}$-parameter diffusion drafter that predicts multiple tokens in parallel, conditioned on target-model features. Thus, \dflash{} introduces roughly three times as many parameters as our method. Its training is also substantially more expensive. For block size $B$ and sequence length $L$, \dflash{} requires a training context length of $B\cdot L$, whereas our method always uses $2\cdot L$, independent of $B$. \textbf{This makes \dflash{} significantly more expensive to train than our method.} We evaluate the open-source checkpoints for EAGLE-3 and \dflash{}; see~\tab{tab:checkpoints}. Refer~\supp{subsec:qwen-sampler-configs} for sampler configurations.

\paragraph{Lossy Baselines (Diffusion Methods)}
We also compare against lossy parallel-generation methods. Jacobi Forcing~\citep{hu2025fast}, SDAR~\citep{cheng2026sdar}, OPDLM~\citep{su2026data}, and I-DLM~\citep{yu2026introspective} are fine-tuned from Qwen3; Fast-dLLM v2~\citep{wu2025fastdllmv2} is fine-tuned from Qwen2.5; FLARE~\citep{zhu2026flare} is fine-tuned from Qwen3.5; and LLaDA2.1-Flash~\citep{bie2026llada21} is trained from scratch. We evaluate the open-source checkpoints for Fast-dLLM v2 and SDAR; see~\tab{tab:checkpoints}. OPDLM, I-DLM, and FLARE are concurrent works.

\paragraph{Metrics} We report average pass@1 accuracy over multiple generations for each benchmark; refer~\tab{tab:benchmark-overview} for details. For speculative decoding methods, one generation step consists of a draft forward pass followed by a verifier forward pass. We report the number of tokens decoded per step, $\tau$, which comprises one draft and one verify step. However, this metric does not account for differences in drafter size and can therefore be misleading, particularly because the baseline drafters are smaller than ours. We consequently measure throughput using the 1K/8K throughput test described earlier. Following standard practice, we do not report accuracy for lossless methods because any differences arise from sampling randomness and numerical nondeterminism. 
We report $\tau$ under sampler configurations optimized for (1) system throughput and (2) per-request throughput. For each method, we select these configurations through a grid search over sampling hyperparameters; see~\tab{tab:all-methods-throughput}.
We use thinking mode for all methods. 
Although \dflash{} recommends disabling it for greater speedups~\citep{zlab2026dflashmodelcard}, doing so reduces average accuracy from $76.36\%$ to $55.40\%$; see~\supp{supp:dflash-thinking}. We therefore retain thinking mode for a fair comparison.
When comparing with lossy methods, we instead report tokens per forward pass (TPF). For our method, $\mathrm{TPF}=\tau/2$ because the drafter and verifier are similar in size.
For both drafting and verification, we use temperature $\temp=1$, top-$p=0.95$, and top-$k=50$ unless otherwise specified. We use a context length of $32{,}768$, the native context length of Qwen3-8B, for all benchmarks and methods.

\subsubsection{\texorpdfstring{\methodqwen{}}{Uno Qwen}: Qwen-based Diffusion Augmented LLM (Ours)}
\paragraph{Autoregressive Weights} We initialize the AR weights from the open-source Qwen3-8B checkpoint~\citep{yang2025qwen3} and keep them frozen during training.

\paragraph{Training Diffusion Weights} For each AR weight matrix, we add a {rank-128} LoRA adapter with $\alpha_{\text{LoRA}}=256$. These adapters introduce $0.35\mathrm{B}$ trainable parameters. We train them for three epochs, corresponding to $14.7\mathrm{B}$ tokens, on OpenThoughts3-1.2M~\citep{guha2025openthoughts}, using a maximum sequence length of $4{,}096$.
We use a block-size curriculum with $B\in\{2,4,6,8,12,16\}$, increasing the block size every half epoch. Training uses a global batch size of $64$ and a constant learning-rate of $10^{-5}$ with 2\% warmup steps. We set the loss coefficients $\alpha=0$ and $\beta=1$.
Training takes approximately 32 hours on 4 nodes, each with 8 H200 GPUs.
For ablations, we train all models for one epoch  with $B=8$, denoted \methodqwenone{}. Unless stated otherwise, we evaluate this variant using the Linear sampler with $B=16$.

\subsubsection{Results}

\paragraph{Comparison with Lossless Speculative Decoding Methods}
\begin{wrapfigure}{r}{0.42\textwidth}
    \centering
    \includegraphics[width=\linewidth]{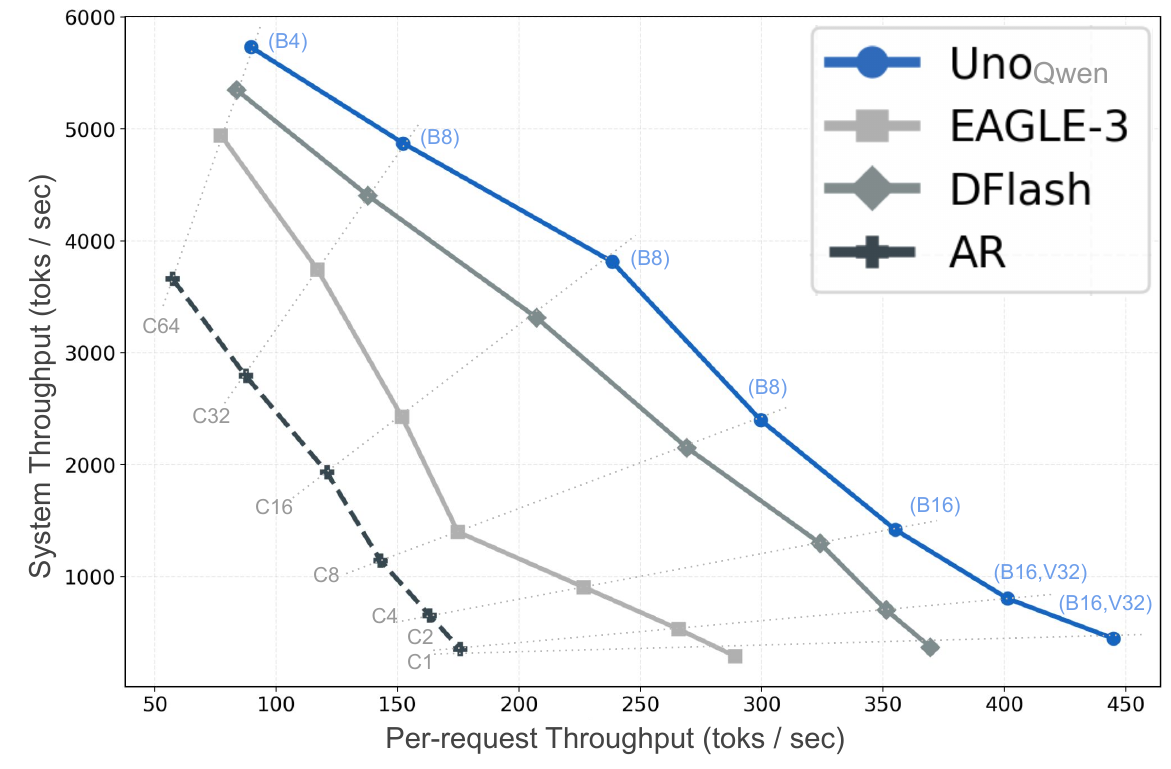}
    \caption{\footnotesize System versus per-request throughput across batch sizes (concurrency $C$). \method{} Pareto-dominates speculative decoding and achieves up to $2.5\times$ speedup over the base AR model. Parentheses indicate the sampler configuration yielding the highest throughput for \method{}. Exact throughputs are in \tab{tab:all-methods-throughput}.}
    \label{fig:uno_spec-dec_pareto}
    \vspace{-1em}
\end{wrapfigure}
\tab{tab:lossless-baselines-tpf-t1} compares the average number of tokens per step, $\tps$, for \methodqwen{}, \eagle{}, and \dflash{} at $\temp=1$. For each method, we identify the configurations that maximize system and per-request throughput through a grid search over sampling parameters, as detailed in~\tab{tab:all-methods-throughput}.
At the largest supported batch size, with Linear sampler $B=4$, all methods achieve their highest system throughput. 
\methodqwen{} exceeds $5700$ tokens per second, outperforming \dflash{} and \eagle{} and achieving a $1.6\times$ speedup over the base AR model. This improvement enables faster RL training and offers practical speedups in LLM inference.
For the best per-request throughput (batch size of 1), \methodqwen{} and \eagle{} perform best with tree-sampler configurations $(B,K,V)=(16,32,32)$ and $(8,32,60)$, respectively, while \dflash{} performs best with $B=16$. \methodqwen{} substantially outperforms both baselines, achieving a $2.5\times$ speedup over the base AR model. It is also Pareto-dominant in throughput across all batch sizes~\fig{fig:uno_spec-dec_pareto}.
Unlike our method, which shares a single KV cache, \eagle{} and \dflash{} maintain separate caches for drafting and verification, resulting in higher peak memory usage.

\begin{remarkbox}
\textbf{Takeaway 4:} {\method{} is strictly faster than \dflash{} and \eagle{} across all batch sizes, with fewer additional parameters and a shared draft-verifier KV cache that reduces peak memory.}
\end{remarkbox}

\paragraph{Comparison with (Lossy) Diffusion Methods}
\tab{tab:speedup-methods} compares \methodqwen{} with the leading lossy diffusion-based acceleration methods for completeness. Unlike our method \methodqwen{}, I-DLM, TiDAR, Jacobi Forcing, FLARE, Fast-dLLM-v2, and SDAR achieve lossy speedups relative to their respective parent models. Despite optimizing for quality first, our method achieves a higher TPF than most methods across numerous benchmarks.  Benchmarks highlighted in red indicate accuracy degradation. FLARE supports both lossy and target-lossless sampling, but the paper does not state which produced the results.

\begin{table*}[t]
\centering
\caption{
Acceptance lengths ($\tau$), throughput (1K/8K test), peak memory usage, and additional parameter counts for \methodqwen{}, \eagle{}, and \dflash{} at sampling $\temp=1$. We report the $\tau$ values that maximize system and per-request throughput.
}
\label{tab:lossless-baselines-tpf-t1}
\resizebox{\textwidth}{!}{%
\begin{tabular}{lZccZcc}
\toprule
& \multicolumn{3}{c}{System Throughput Optimal}
& \multicolumn{3}{c}{Per-request Throughput Optimal} \\
\cmidrule(lr){2-4}\cmidrule(lr){5-7}
& \normalZ{\shortstack{\methodqwenc{}\\$B$=$4$}}
& \shortstack{EAGLE-3\\$B$=$4$}
& \shortstack{DFlash\\$B$=$4$}
& \normalZ{\shortstack{\color{zaibluefg} \methodqwenc{}\\$B$=$16$, $V$=$32$}}
& \shortstack{EAGLE-3\\$B$=$8$, $V$=$60$}
& \shortstack{DFlash\\$B$=$16$} \\
\midrule
\textit{Math} & & & & & & \\
\quad GSM8K        & \textbf{3.99} & 2.26 & 2.47 & \textbf{6.29} & 3.83 & 3.38 \\
\quad MATH500      & \textbf{4.11} & 2.14 & 2.28 & \textbf{6.89} & 3.52 & 3.24 \\
\quad AIME-24      & \textbf{4.08} & 2.11 & 1.96 & \textbf{6.83} & 3.41 & 2.76 \\
\quad AIME-25      & \textbf{4.11} & 2.13 & 1.90 & \textbf{6.81} & 3.46 & 2.56 \\
\quad AIME-26      & \textbf{4.08} & 2.13 & 1.93 & \textbf{6.71} & 3.49 & 2.57 \\
\midrule
\textit{Coding} & & & & & & \\
\quad HumanEval    & \textbf{3.83} & 2.12 & 2.35 & \textbf{5.63} & 3.71 & 3.03 \\
\quad MBPP         & \textbf{3.87} & 2.16 & 2.22 & \textbf{5.87} & 3.65 & 3.04 \\
\quad LCBv6        & \textbf{3.77} & 2.04 & 1.76 & \textbf{5.34} & 3.45 & 2.22 \\
\midrule
\textit{Science and Knowledge} & & & & & & \\
\quad GPQA         & \textbf{3.79} & 1.99 & 1.98 & \textbf{5.49} & 3.25 & 2.57 \\
\quad GPQA-Diamond & \textbf{3.78} & 1.98 & 1.92 & \textbf{5.49} & 3.19 & 2.50 \\
\quad MMLU-Pro     & \textbf{3.84} & 2.03 & 2.11 & \textbf{5.70} & 3.31 & 2.80 \\
\midrule
\textit{Instruction Following} & & & & & & \\
\quad IFEval       & \textbf{3.48} & 1.91 & 1.92 & \textbf{4.58} & 3.49 & 2.26 \\
\midrule
{$\tps$} & \textbf{3.89} & 2.08 & 2.07 & \textbf{5.97} & 3.48 & 2.74 \\
{Throughput (Toks / sec; $\uparrow$)} & \textbf{5733} & 4944 & 5351 & \textbf{445} & 284 & 370 \\
{Peak Memory (GiB; $\downarrow$)} & \textbf{122.2} & 130.0 & 130.1 & \textbf{118.0} & 129.4 & 129.8 \\
{Additional Params (B; $\downarrow$)} & \textbf{0.35} & 0.40 & 1.05 & \textbf{0.35} & 0.40 & 1.05 \\
\bottomrule
\end{tabular}
}

\end{table*}

\subsubsection{Ablation}\label{sec:experiments:ablation}
We ablate the main training choices in \tab{tab:qwen-owt-ablation} using \methodqwenone{} with the Linear sampler and $B=16$.

\paragraph{Ablation 1: Loss Terms} 
The diffusion loss in \Eqn{eqn:training-loss} combines the distillation loss, $\mathcal{L}_{\text{DCD}}$, and the total variation loss, $\mathcal{L}_{\text{TV}}$. Training with $\mathcal{L}_{\text{TV}}$ alone achieves a TPF of $2.39$, outperforming both $\mathcal{L}_{\text{DCD}}+\mathcal{L}_{\text{TV}}$ ($2.23$) and $\mathcal{L}_{\text{DCD}}$ alone ($2.23$). Further analysis revealed that the $\mathcal{L}_{\text{DCD}}$ loss was an order of magnitude larger than the $\mathcal{L}_{\text{TV}}$ loss. Consequently, reducing the $\mathcal{L}_{\text{DCD}}$ weight to $0.01$ yielded a marginal improvement, increasing the TPF to $2.40$.

\paragraph{Ablation 2: Training Curriculum} In~\supp{supp:block-curricula}, we show that 
increasing the training block size from 4 to 16 every half epoch improves TPF from $2.65$ to $2.71$, compared with using a fixed block size of 16 for two epochs.

\paragraph{Ablation 3: Diffusion Weights Configuration} 
In~\supp{supp:lora-adapter}, we show that distributing diffusion adapters across all layers is more effective than placing the same number of parameters in only a subset of layers. We also ablate the LoRA rank $r_{\text{LoRA}}$ and scale $\alpha_{\text{LoRA}}$. Increasing $r_{\text{LoRA}}$ from $128$ to $256$ improves TPF, as shown in~\tab{tab:qwen-owt-ablation}, but also increases inference cost. Across the tested values of $\alpha_{\text{LoRA}}/r_{\text{LoRA}}$, a ratio of $64$ performs best.

\begin{table*}[t]
\centering
\caption{
Benchmark accuracy (ACC) and TPF for our {\color{zaibluefg} lossless} method, \methodqwenc{}, and {\color[RGB]{250,142,142} lossy} diffusion methods. Entries are reported as $\mathrm{ACC}_{\mathrm{TPF}}$. 
For \methodqwenc{}, TPF uses $\temp=0$ and the tree sampler with $(B,K,V)=(16,32,60)$. Lossy speedups are taken from the respective papers, except for Fast-dLLM v2 and SDAR, which we evaluated using the provided checkpoints. Accuracy drops relative to the corresponding parent AR model are shaded in \newnewourredsample{red} by severity. Jacobi Forcing\textsuperscript{J}  uses separate math and coding models trained from Qwen2.5-Math-7B-Instruct and Qwen2.5-Coder-7B-Instruct, respectively. \textsuperscript{F} Fast-dLLM v2 uses Qwen2.5-7B-Instruct.
}
\label{tab:speedup-methods}
{
\resizebox{\textwidth}{!}{%
\setlength{\tabcolsep}{3pt}
\begin{tabular}{lZcccccccc}
\toprule
&\normalZ{ } & \multicolumn{8}{c}{{\color[RGB]{250,142,142} Lossy} Speedup Methods} \\
\cmidrule(lr){3-10}
Benchmark
& \normalZ{{\color{zaibluefg} \methodqwen{}}}
& SDAR
& \shortstack{TiDAR\\(Trust Diff)}
& OPDLM
& I-DLM
& \shortstack{Jacobi\\Forcing (MR)\textsuperscript{J}}
& \shortstack{Fast-\\dLLM v2\textsuperscript{F}}
& \shortstack{FLARE\\(AR Trust)\textsuperscript{L}}
& \shortstack{LLaDA2.1-\\Flash (S Mode)} \\
\textit{Model size}
& \normalZ{$8\mathrm{B}$}
& $8\mathrm{B}$
& $8\mathrm{B}$
& $8\mathrm{B}$
& $8\mathrm{B}$
& $7\mathrm{B}$
& $7\mathrm{B}$
& $9\mathrm{B}$
& \shortstack{$100\mathrm{B}$-$\mathrm{A}5\mathrm{B}$} \\
\textit{Parent AR}
& \multicolumn{5}{c}{Qwen3-8B}
& \multicolumn{2}{c}{Qwen2.5-7B Family}
& Qwen3.5-9B
& \na{} \\

\cmidrule(lr){1-1}
\cmidrule(lr){2-6}
\cmidrule(lr){7-8}
\cmidrule(lr){9-9}
\cmidrule(lr){10-10}

\textit{Math} & & & & & & & & & \\

\quad GSM8K
& $96.1_{3.56}$
& \newnewourredcell{$91.4_{2.3}$}
& \newnewourredcell{$80.4_{7.1}$}
& \newnewourredcell{$87.1_{1}$}
& \newnewourredcellgrad{55}{$95.0_{\na{}}$}
& \newnewourredcellgrad{50}{$91.4_{4.0}$}
& $83.7_{3.0}$
& $93.3_{\na{}}$
& -- \\

\quad MATH500
& $96.4_{3.92}$
& \newnewourredcell{$72.0_{2.8}$}
& --
& \newnewourredcell{$71.2_{1}$}
& $96.8_{\na{}}$
& --
& \newnewourredcell{$61.1_{2.1}$}
& \newnewourredcellgrad{70}{$95.2_{\na{}}$}
& -- \\

\quad AIME-24
& $76.7_{4.01}$
& \newnewourredcell{$10.0_{2.8}$}
& --
& \newnewourredcell{$14.7_{1}$}
& \newnewourredcell{$69.6_{\na{}}$}
& --
& $6.67_{2.7}$
& \newnewourredcell{$63.3_{\na{}}$}
& -- \\

\quad AIME-25
& $76.7_{3.96}$
& \newnewourredcell{$10.0_{2.6}$}
& --
& \newnewourredcell{$12.4_{1}$}
& \newnewourredcell{$60.8_{\na{}}$}
& --
& $0.0_{2.6}$
& \newnewourredcell{$54.4_{\na{}}$}
& $63.3_{5.4}$ \\

\quad AIME-26
& $73.3_{4.05}$
& --
& --
& --
& --
& --
& --
& --
& -- \\

\textit{Coding} & & & & & & & & & \\

\quad HumanEval
& $94.8_{3.67}$
& \newnewourredcell{$75.6_{2.8}$}
& \newnewourredcell{$57.9_{7.3}$}
& \newnewourredcell{$59.8_{1}$}
& \newnewourredcellgrad{75}{$93.3_{2.6}$}
& \newnewourredcell{$83.5_{4.1}$}
& $63.4_{2.5}$
& \newnewourredcell{$92.1_{\na{}}$}
& -- \\

\quad MBPP
& $89.0_{4.21}$
& \newnewourredcell{$68.1_{1.5}$}
& \newnewourredcell{$65.4_{10.0}$}
& \newnewourredcell{$48.7_{1}$}
& $92.2_{\na{}}$
& \newnewourredcell{$70.4_{2.8}$}
& $63.0_{4.5}$
& $91.1_{\na{}}$
& -- \\

\quad LCBv6
& $51.4_{3.75}$
& \newnewourredcell{$16.6$}
& --
& \newnewourredcell{$9.7_{1}$}
& \newnewourredcell{$45.7_{\na{}}$}
& --
& $10.0_{1.7}$
& $49.7_{\na{}}$
& $44.1_{6.5}$ \\

\multicolumn{2}{l}{\textit{Science and Knowledge}} & & & & & & & & \\

\quad GPQA
& $58.0_{4.10}$
& --
& --
& --
& \newnewourredcell{$54.9_{\na{}}$}
& --
& \newnewourredcellgrad{80}{$31.9_{\na{}}$}
& --
& -- \\

\quad GPQA-D
& $60.9_{4.01}$
& \newnewourredcell{$40.2$}
& --
& \newnewourredcell{$36.1_{1}$}
& \newnewourredcell{$55.6_{\na{}}$}
& --
& $21.7_{2.2}$
& \newnewourredcell{$71.2_{\na{}}$}
& $66.7_{4.0}$ \\

\quad MMLU-Pro
& $74.8_{3.61}$
& \newnewourredcell{$56.9_{1.6}$}
& --
& \newnewourredcell{$53.7_{1}$}
& \newnewourredcellgrad{85}{$73.1_{\na{}}$}
& --
& --
& \newnewourredcell{$77.4_{\na{}}$}
& $75.3_{4.4}$ \\

\multicolumn{2}{l}{\textit{Instruction Following}} & & & & & & & & \\

\quad IFEval
& $86.5_{2.49}$
& \newnewourredcell{$61.4_{1.5}$}
& --
& \newnewourredcell{$50.1_{1}$}
& \newnewourredcellgrad{90}{$84.7_{\na{}}$}
& --
& \newnewourredcell{$61.4_{1.5}$}
& \newnewourredcell{$71.4_{\na{}}$}
& $83.4_{2.2}$ \\

\bottomrule
\end{tabular}%
}
}
\end{table*}

\section{Related Work}
\paragraph{Speculative Decoding} 
Speculative decoding methods such as \eagle{} and \dflash{} also provide lossless speedups but require training a separate draft model that is smaller than the target model. Designing a draft model requires numerous choices about its depth, hidden dimension, MLP expansion, attention heads, parameter sharing, and KV projections for target features.  \citet{li2026diffuspec,sandler2026specdiff} on the other hand use large pretrained dLLMs for drafting and separate AR models for verification. These large drafters incur substantial memory and latency overhead, making these methods slower than \dflash{}. In contrast, \method{} uses a single architecture with distinct drafting and verification pathways. Consequently, \method{} maintains a single KV cache, introduces fewer additional parameters, and requires less peak inference memory than conventional speculative decoding. These advantages yield larger speedups at high batch sizes, making \method{} a drop-in replacement for methods that use separate draft and target models.

\paragraph{Self-Speculative Decoding} 
\citet{sahoo2025esoteric} introduced a hybrid AR-diffusion framework in which a single denoising model performs either multi-token diffusion generation or AR generation, depending on the available compute budget. This approach substantially outperforms block diffusion~\citep{arriola2025block}. Building on self-speculative decoding~\citep{Stern2018BlockwisePD}, \citet{liu2025tidar} extended this framework by using the same denoising model to draft sequences in diffusion mode and verify them via rejection sampling in AR mode, yielding significant qualitative improvements over standalone diffusion generation. However, because this method modifies the base AR model's weights to support diffusion generation, it does not preserve the model's original distribution. Moreover, its speedups are limited to small batch sizes~\citep{fu2026nemotronlabsdiffusion}. In contrast, \method{} preserves the base AR model's distribution while providing lossless speedups, making it a drop-in alternative to these methods.

\paragraph{Diffusion Models} 
Large-scale d-LLMs~\citep{nie2025large,gat2025set,bie2026llada21,fu2026nemotronlabsdiffusion,diffusiongemma} can generate faster than similarly sized AR models at small batch sizes but generally lag behind them in quality. Their lossy speedups also diminish at larger batch sizes, limiting their applicability. Post-training these models is more expensive because rollouts become slower at large batch sizes and existing RL recipes and objectives require nontrivial modifications~\citep{zhao2026d1,wang2025d2}. In contrast, \method{} provides lossless speedups across batch sizes, uses standard AR post-training algorithms without modification, and substantially accelerates post-training.

\paragraph{Multi-Token Prediction (MTP)}  
Methods such as Medusa~\citep{cai2024medusa} and the approach of \citet{gloeckle2024better} provide lossless speedups but modify the transformer architecture to predict future tokens. \citet{li2026eagle} showed that speculative decoding with a separately trained draft model is faster than these approaches. We further show that \method{} outperforms speculative decoding methods and, consequently, these MTP methods. Nevertheless, architectural changes for predicting future tokens are complementary to \method{} and could improve its draft acceptance rate. We leave this combination to future work.

\paragraph{Quadratic Samplers} 
Each generation step in our sampler uses two forward passes: one for drafting and one for verification. Quadratic sampling, proposed by \citet{samragh2025your} and used in TiDAR and \nld{}, can combine these operations into a single pass. For a draft of $k$ tokens, it inserts $k$ masked placeholders after each draft position, yielding $k^2$ masked positions that represent possible future continuations. This structure verifies the current draft while generating candidates for the next step. Practical speedups require efficient kernels, which we leave to future work.

\paragraph{Comparison with Concurrent work} 
Concurrent methods such as OPDLM, FLARE, and I-DLM provide lossy speedups relative to the base AR model because they fine-tune its weights; see \tab{tab:speedup-methods}. 
Although I-DLM is lossy, its I-DLM (R-ISD) variant uses gated LoRA. Despite the authors' theoretical argument that its sampler is lossless, we could not reproduce lossless speedups using the released LoRA adapters and sampler implementation. Because this method is compatible with our proposed $\Psi$-spec sampler using a mask prior, we evaluate the combination and show that it achieves lossless speedups, although it remains substantially slower than our method. We provide a detailed comparison with I-DLM, including an evaluation of its official LoRA checkpoint, in \supp{supp:I-DLM}.

\section{Conclusion}
In this work, we introduced diffusion-augmented LLMs, which unify two distinct generation pathways within a single architecture: an autoregressive pathway parameterized by $\wbase$ and a diffusion pathway parameterized by $\wbase + \wlora$. We showed that training the diffusion weights requires orders of magnitude fewer tokens than the AR weights. This formulation allows \method{} either to be trained from scratch or to be created by augmenting an existing AR model with lightweight diffusion weights for faster generation.

\method{} serves as a drop-in alternative to speculative diffusion methods such as \eagle{} and \dflash{}, without requiring a separate draft model, and to self-speculative methods such as TiDAR, without introducing lossy speedups. Across the evaluated batch sizes, \method{} achieves up to a $3\times$ speedup over the base AR model, while retaining up to a $2\times$ speedup at the largest batch size supported by that model. These gains accelerate both inference and RL post-training, where rollout generation is a major computational bottleneck. More broadly, diffusion-augmented LLMs expand the design space of AR language models by combining the quality of AR generation with the parallelism of diffusion in a unified architecture.

\paragraph{Acknowledgement} We thank A. Feder Cooper, Willis Guo, Rupesh Srivastava, and Matthew Yang for their helpful feedback and insightful discussions.

\bibliography{appendix}
\bibliographystyle{appendix}

\appendix
\clearpage
\setcounter{tocdepth}{2}
\tableofcontents
\section{Background}\label{supp:background}
\subsection{Masked Diffusion}\label{supp:background:masked_diffusion}
MDMs use a masked prior, where $\prior = \m \in \onehotset$ is the one-hot representation of a special \mask token. During the forward process~\Eqn{eqn:interpolating-forward}, tokens either remain unchanged or transition to the masked state  $\m$, after which they stay masked. This behavior carries over to the reverse process. The posterior of the reverse process $q^{\text{MDM}}_{s|t}$ for $0 \leq s < t < 1$ can be derived using Bayes' Rule, and is given by:
\begin{align}
    \label{eqn:mdm-posterior}
    [q^\ell_{s|t}(\cdot\mid \z_t, \x)]^{\text{MDM}} = 
    \begin{cases}
    \cat\left(\ \cdot\ ;\  \frac{\alpha_s - \alpha_t}{1 - \alpha_t} \x^{\supl} + \frac{1 - \alpha_s}{1 - \alpha_t}\ztl \right) & \text{if $\ztl = \m$,} \\
    \cat(\ \cdot\ ;\  \x^{\supl}) & \text{otherwise.}
    \end{cases}
\end{align}
The learned reverse posterior is $$[\pst(\z_s \mid \z_t)]^{\text{MDM}} =  q^{\text{MDM}}_{s|t}(\z_s \mid \z_t, \x = \x_\theta(\z_t, t)),$$ where $\denoise: \onehotset ^ L \times [0, 1]\to \Delta^{L}$ is the denoising model.
A key limitation is that once unmasked, tokens cannot be remasked. This can create compounding errors during inference, as the denoising model $\denoise$ imperfectly models the clean data.

\paragraph{NELBO}
The likelihood of a sequence under the learned reverse diffusion process is generally intractable to evaluate exactly, as it requires marginalizing over all possible unmasking trajectories. Therefore, typically the negative evidence lower bound (NELBO) is minimized, which is a tractable upper bound on the negative log-likelihood. For continuous-time masked diffusion, the NELBO simplifies to the weighted denoising objective \citep{sahoo2024simple, shi2024simplified, ou2025your}
\begin{align}\label{eqn:mdlm_nelbo}
\mathcal{L}_{\text{NELBO}}^{\text{MDM}}
= \mathbb{E}_{t\sim \mathcal{U}[0, 1],q_t} \frac{\dat}{1 - \at} \log \langle \x^\ell_\theta(\z_t, t), \x^\ell\rangle.
\end{align}

\subsection{Uniform-State Diffusion Models}\label{supp:background:uniform_diffusion}
We now focus on Uniform-State Diffusion Models (USDMs), which use a uniform prior $\prior=\vone/K$~\citep{austin2021structured,sahoo2025the}. Unlike masked diffusion, where tokens are fixed once unmasked, USDMs allow continuous updates of all token positions, which naturally supports self-correction at inference. This capability makes them particularly effective for few-step generation~\citep{sahoo2025the}, inference-time scaling~\citep{deschenaux2026the} and guided sampling~\citep{schiff2025simple}.
Given a noisy sequence $\z_t$ at time $t$, the Markov property defines the reverse posterior $q_{s|t}$ for obtaining a less noisy sequence $\z_s$ ($0 \leq s < t$) by sampling from the following distribution:
{\small
\begin{equation}\label{eqn:uniform_true_reverse_posterior}
    [{q^{{\supl}}_{s|t}}(\cdot \mid \z_t, \x)]^{\text{USDM}} = \cat\left(\cdot\ ;\   \frac{K\at \ztl \odot \xl + ({ \alpha_{t|s}} - \at)\ztl + (\alpha_s - \alpha_t)\xl + (1 - {\alpha_{t|s}})(1- \as)\vone / K}{K \at\langle \ztl, \xl\rangle  + 1 - \at}\right).
\end{equation}
}
Similar to the posterior for masked diffusion models above, the true reverse posterior~\Eqn{eqn:uniform_true_reverse_posterior} depends on the unknown clean sequence $\x$, hence it is approximated as:
\begin{equation}
    [\pst(\z_s \mid \z_t)]^{\text{USDM}} =  q^{\text{USDM}}_{s|t}(\z_s \mid \z_t, \x = \x_\theta(\z_t, t)),
\end{equation}
where $\denoise:\onehotset^L\times [0, 1] \rightarrow \Delta^L$ is the denoising network parameterized by $\theta$.

\paragraph{NELBO}
This posterior induces the following NELBO~\citep{sahoo2025the}:
\begin{align}\label{eqn:loss-usdm-singleline}
    \mathcal{L}_{\text{NELBO}}^{\text{USDM}}\left({ q}, p_\theta; \x \right) & = -\mathbb{E}_{t \sim \mathcal{U}[0,1],q_t} 
    \;\sum_{\ell \in [L]}\felbo(\ztl, \denoise^{\supl} (\ztl, t), \at; \xl),
\end{align}
where 
{\footnotesize
\begin{align}
    \felbo(\ztl, \denoise^{\supl}(\ztl, t), & \at; \xl) = \frac{\at'}{K\at}\Bigg[\frac{K}{\barx^{\supl}_r} - \frac{K}{(\barx_\theta^{\supl})_r} \nonumber - \left(\zeta_t \mathds{1}_{\ztl = \xl} + \mathds{1}_{\ztl \neq \xl}\right)\sum_{j} \log\frac{(\barx^{\supl}_\theta)_r}{(\barx^{\supl}_\theta)_j} \\
    & - K\frac{\at}{1 - \at} \log \frac{(\barx^{\supl}_\theta)_r}{(\barx^{\supl}_\theta)_i}\mathds{1}_{\ztl \neq \xl} - \left((K - 1)\zeta_t \mathds{1}_{\ztl = \xl} - \frac{1}{\zeta_t}\mathds{1}_{\ztl \neq \xl}\right)\log \zeta_t
    \Bigg].
\end{align}
}
Here, $\barx^{\supl} = K\alpha_t\x^{\supl} + (1 - \alpha_t)\vone$, $\barx^{\supl}_\theta = K\alpha_t \denoise^{\supl}(\z_t, t) + (1 - \alpha_t)\vone$, $\alpha_t'$ denotes the time derivative of $\alpha_t$, $r = \arg\max_{j\in [K]} (\z^{\supl}_t)_j$ is the nonzero entry of $\z_t$, $\zeta_t = \frac{1 - \alpha_t}{K \alpha_t + 1 - \alpha_t}$, and $i$ denotes the index in $\x$ corresponding to 1, that is, $\x_i = 1$.

\subsection{Self-speculative Decoding}
Self-speculative decoding eliminates the need for a separate draft model by using a single network for both proposal and verification. \citet{Stern2018BlockwisePD} first proposed augmenting the base model with auxiliary prediction heads to predict a block of future tokens in parallel, then using the same model to verify these predictions and commit the longest validated prefix. Their exact-match verification rule guarantees equivalence to greedy decoding, rather than exact sampling. \citet{liu2025tidar} instead convert an AR model into a diffusion model: the model drafts tokens in diffusion mode and verifies them in AR mode, with the attention pattern changing between the two modes while the model weights remain the same in both modes. They combine this approach with the rejection-sampling acceptance rule of \citet{leviathan2023fast} and \citet{chen2023accelerating} to recover exact sampling from the base model's distribution. Concurrent work \citep{zhu2026flare, yu2026introspective} builds on the same idea. \textbf{Crucially, however, this approach requires modifying the base AR model's weights, so it is not lossless.}

\subsection{\texorpdfstring{$\Psi$}{Psi}-Samplers}\label{supp:background:psi-samplers}
$\Psi$-Samplers~\citep{deschenaux2026the} are predictor-corrector samplers for discrete diffusion models that share the marginals of the ancestral sampler. They interpolate between the reverse posterior and the forward process~\Eqn{eqn:interpolating-forward}: 
\begin{align}\label{eqn:psi}
\Psi_{s\mid t}(\cdot\mid \x,\z_t; \prior)
= \kappa_t {q}_{s\mid t}(\cdot\mid \z_t,\x; \prior)
+ (1-\kappa_t)q_s(\cdot\mid \x; \prior),
\end{align}
where $\prior$ is the noise prior (e.g., the masked prior $\prior=\m$ for MDMs or the uniform prior $\prior=\vone/K$ for USDMs), ${q}_{s\mid t}(\cdot \mid \cdot\ ;\   \prior)$ is the reverse posterior for MDMs \Eqn{eqn:mdm-posterior} or USDMs \Eqn{eqn:uniform_true_reverse_posterior}, $q_s$ is the forward noising process, and $\kappa_t\in[0,1]$ sets the correction strength. The first term \textit{predicts} by denoising the current state; the second \textit{corrects} by re-injecting noise, so earlier decisions can be revised. Setting $\kappa_t=1$ recovers ancestral sampling.

Since $\x$ is unavailable at generation time, the practical sampler substitutes the denoiser prediction $\denoise(\z_t,t)$, giving the token-wise transition
\begin{align}\label{eqn:psi-sampler}
    & [\Psi^\theta_{s | t}(\cdot\mid \z_t; \denoise(\z_t, t),  \prior)]^{\supl} \nonumber \\
    & = \kappa_t {q_{s|t}^{{\supl}}}(\cdot\mid  \z_t,  \denoise(\z_t, t);  \prior) + (1 - \kappa_t) 
    \left[
    \alpha_s {q_{0|t}^{{\supl}}}(\cdot\mid  \z_t, \denoise(\z_t, t);  \prior) + (1 - \alpha_s)  \prior \right].
\end{align}
Ancestral samplers can lock in incorrect early predictions, whereas $\Psi$-samplers correct errors explicitly: for MDMs the corrector can remask decoded tokens, and for USDMs it assigns every token non-zero probability, permitting transitions into states that the denoiser incorrectly scores as low-probability.
Sample quality therefore keeps improving with additional steps.

For MDMs, the transition \Eqn{eqn:psi-sampler} simplifies to
\begin{align}
&[\Psi^\theta_{s\mid t}(\cdot\mid\z_t;\denoise(\z_t, t),\prior = \m)]^{\supl}
\nonumber\\
&=
\begin{cases}
\operatorname{Cat}\!\left(
\cdot\,;\,
[\alpha_s+\kappa_t(1-\alpha_s)]
\x_\theta^\supl(\z_t,t)
+
(1-\kappa_t)(1-\alpha_s)\m
\right),
&
\z_t^{\supl}\neq\m,
\\[2mm]
\operatorname{Cat}\!\left(
\cdot\,;\,
\left[
\alpha_s-\frac{\kappa_t\alpha_t(1-\alpha_s)}{1-\alpha_t}
\right]
\x^{\supl}_\theta(\z_t, t)
+
\left[
1-\alpha_s+\frac{\kappa_t\alpha_t(1-\alpha_s)}{1-\alpha_t}
\right]\m
\right),
&
\z_t^{\supl}=\m.
\end{cases}
\label{eqn:psi-mdm}
\end{align}
To obtain the marginals for single-step generation, we simply plug in $s=0$ and $t=1$ into \Eqn{eqn:psi-mdm} which gives:
\begin{align}
&[\Psi^\theta_{s=0\mid t=1}(\cdot\mid\z_t;\denoise(\z_t, t),\prior = \m)]^{\supl}
\nonumber\\
&=
\begin{cases}
\operatorname{Cat}\!\left(
\cdot\,;\,
\left[1+\kappa_t(1-1)\right]
\x^{\supl}_\theta(\z_1, 1)
+
(1-\kappa_t)(1-1)\m
\right),
&
\z_1^{\supl}\neq\m,
\\[2mm]
\operatorname{Cat}\!\left(
\cdot\,;\,
\left[
\alpha_s-\frac{\kappa_t\cdot 0 \cdot(1-1)}{1-0}
\right]
\x^{\supl}_\theta(\z_1, 1)
+
\left[
1-1+\frac{\kappa_t\cdot 0 \cdot (1-1)}{1-0}
\right]\m
\right),
&
\z_1^{\supl}=\m.
\end{cases} \\
&=
\begin{cases}
\cat(\cdot; \x_\theta^\ell(\z_1)),
&
\z_1^{\supl}\neq\m,
\\[2mm]
\cat(\cdot; \x_\theta^\ell(\z_1)),
&
\z_1^{\supl}=\m.
\end{cases} \\
& = \cat(\cdot; \x_\theta^\ell(\z_1))
\label{eqn:psi-mdm-single-step} 
\end{align}

For USDM, the transition \Eqn{eqn:psi-sampler} simplifies to:
\begin{align}
&[\Psi^\theta_{s\mid t}
(\cdot\mid\z_t;\denoise(\z_t, t), \prior = \vone/K)]^{\supl}
\nonumber\\
&=
\operatorname{Cat}\Bigg(
\cdot\,;\,
\frac{1}{
K\alpha_t
\left\langle
\z_t^{\supl},[\denoise(\z_t, t)]^{\supl}
\right\rangle
+1-\alpha_t
}
\Bigg[
K\alpha_t\left[\alpha_s+\kappa_t(1-\alpha_s)\right]
\z_t^{\supl}\odot[\denoise(\z_t, t)]^{\supl}
\nonumber\\
&\qquad\qquad
+\kappa_t(1-\alpha_s)\alpha_{t\mid s}\z_t^{\supl}
+
\left[
\alpha_s(1-\alpha_t)
-\kappa_t\alpha_t(1-\alpha_s)
\right]
[\denoise(\z_t, t)]^{\supl}
\nonumber\\
&\qquad\qquad
+(1-\alpha_s)
\left[
\kappa_t(1-\alpha_{t\mid s})
+
(1-\kappa_t)
\left(
K\alpha_t
\left\langle
\z_t^{\supl},[\denoise(\z_t, t)]^{\supl}
\right\rangle
+1-\alpha_t
\right)
\right]
\frac{\vone}{K}
\Bigg]
\Bigg).
\label{eqn:psi-usdm}
\end{align}
To obtain the marginals for single-step generation, we simply plug in $s=0$ and $t=1$ into \Eqn{eqn:uniform_true_reverse_posterior} which gives
\begin{align}
[\Psi^\theta_{s=0\mid t=1}
(\cdot\mid\z_t;\denoise(\z_t), \prior = \vone/K)]^{\supl} = \cat (\cdot; \x^{\supl}_\theta(\z_1, 1)).
\label{eqn:psi-usdm-single-step} 
\end{align}
\subsection{Discrete Consistency Distillation}\label{supp:background:dcd}
\citet{sahoo2025the} show that Uniform-state discrete diffusion emerges from an
underlying Gaussian diffusion process~\citep{sohl2015deep, song2020score,
kingma2021variational} defined on the one-hot representation $\xl \in \onehotset$:
applying an $\argmax$ projection to the Gaussian latents recovers the discrete
process. This correspondence lets us transfer techniques from the continuous to
the discrete domain, most notably distillation for few-step generation.
Distillation requires deterministic trajectories, namely the probability-flow ODE,
which exist for Gaussian but not for discrete diffusion. Since discrete latents are
simply $\argmax$ projections of Gaussian ones, \citet{sahoo2025the} construct the
trajectory in continuous space and map it back to the discrete domain.

Discrete Consistency Distillation (DCD) trains few-step generators on these
trajectories. First, the Gaussian probability-flow ODE is constructed on the one-hot vectors
and the resulting continuous states are mapped back to tokens via $\argmax$,
yielding a deterministic sequence of discrete latents. The teacher $\teacher:\onehotset^L\times [0, 1] \rightarrow \Delta^L$ is a
neural network with parameters $\theta_0$ that are held fixed, while the student
$\student:\onehotset^L\times [0, 1] \rightarrow \Delta^L$ is a neural network whose parameters $\theta$ are trained. Given two
adjacent states along this path, a noisier $\z_t$ at time $t$ and a less noisy
$\z_s$ at time $s = t - T$, the
student takes the noisier state as input and is trained to match the teacher's
clean-token distribution at the less noisy one, in the spirit of consistency
distillation~\citep{song2023consistencymodels, sahoo2025the}:
\begin{align}
\mathcal{L}_{\mathrm{DCD}}(\studentw;\teacherw)
=
\sum_{\ell=1}^{L}
D_{\mathrm{KL}}\!\left(
\student^\ell(\ztL,t)\,\middle\|\, \teacher^\ell(\ztL,\ s)
\right),
\end{align}
Note that the teacher's parameters $\teacherw$ are not updated during training.
Distillation proceeds over numerous rounds, with the step size $\Delta$ increased
at every round; at the end of each round the trained student parameters $\theta$
are copied into the teacher, which then serves as the target for the next round.
By taking progressively larger denoising steps while remaining consistent along a
shared trajectory, DCD compresses a many-step USDM sampler into a few-step one,
making Uniform-state diffusion particularly effective in the low-step regime.

\paragraph{Limitation:} DCD is trained on deterministic probability-flow
trajectories, which differ from the stochastic denoising trajectories followed at
sampling time. This train-test mismatch limits its effectiveness.

\section{\texorpdfstring{$\Psi$}{Psi}-{Spec} samplers}
\subsection{Diffusion Proposal Distribution Extended}\label{supp:samplers:draft:proposal}
Given a noisy block $\z_t$, we use the $\Psi$-Spec transition in \Eqn{eqn:psi-sampler} to sample a less noisy block $\z_s$, where $s<t$. Because the denoiser uses next-token-prediction (NTP) parameterization, the clean-token distributions used to construct $\z_s$ are given by $\x^{1:B-1}_{\wbase, \wlora}([\clean, \z_t])$. The resulting transition is obtained by drawing each token in parallel from its corresponding marginal distribution:
\begin{numcases}{\z_s^{\ell} \sim}
    \Psi^{\ell}_{s|t} \left(\cdot \mid \z_t ; \x^{{\color{grayseq} 1:B-1}}_{\wbase, \wlora}(\cdot),   {\color{grayseq} \prior{}} \right), \; & $\ell > 1$ \label{eqn:psispec-diffusion-remaining}\\
    \Psi^{\ell=1}_{s|t} \left(\cdot \mid \z_t ; \x^{{\color{grayseq} 1:B-1}}_{\wbase}(\cdot),   {\color{grayseq} \prior{}} \right). \; & $\ell = 1$ \label{eqn:psispec-diffusion-one}
\end{numcases}
\textbf{Notice that the logits for the first, clean position use only the base AR parameters $\wbase$, as shown in \Eqn{eqn:psispec-diffusion-remaining}, whereas the logits at the noisy positions use both the AR weights and the diffusion adapters}. This separation prevents distribution shift because the diffusion adapters are trained only to denoise noisy tokens.  We compute \Eqn{eqn:psispec-diffusion-remaining} and \Eqn{eqn:psispec-diffusion-one} in a single forward pass using the gated LoRA technique of \citet{samragh2025your}.

\subsection{Single-Step Generation}\label{supp:subsec:singlestep}
After removing time conditioning from the denoising model and shifting the logits one position to the left to account for its next-token prediction (NTP) formulation in \Eqn{eqn:psi-mdm-single-step} for MDMs and \Eqn{eqn:psi-usdm-single-step} for USDMs, we show that the resulting marginal distribution has the following form. Let $\Psi_0$ denote the distribution induced by the $\Psi$-sampler at $t=0$: 
\begin{align}
    \Psi_0^{\ell} = 
    \begin{cases}
        \x^{\ell}_{\wbase, \wlora}([\clean, \z_t]), & \ell > 1 \\
        \x^{\ell}_{\wbase}([\clean, \z_t]). & \ell = 1  
    \end{cases}
\end{align}

\subsection{Sampling Algorithm}\label{supp:subsec:sampling-algo}

\alg{alg:psi-spec} shows \method{}'s complete sampling algorithm for the single-sequence case. Differences to speculative decoding~\citep{leviathan2023fast} are written in \UnoDiff{orange}.

\begin{algorithm}[h]
\caption{Lossless $\Psi$-Spec decoding for USDMs (Linear Sampler)}
\label{alg:psi-spec}
\begin{algorithmic}[1]
\Require Prefix $\x$ of length $L$, target model $\x_{\wbase, \wlora}$, block size $B$
\State {Sample noise
    $\z \overset{\mathrm{i.i.d.}}{\sim} \prod_1^{B-1}\mathcal{U}[\mathcal{V}]$}  \Comment{Initialize a block of $B-1$ random tokens}
\State \UnoDiff{$[\mathbf{q}_0,\mathbf{q}] 
    \gets \x_{\wbase,\wlora} ([\x^L,\z]; \x^{<L}) $}
    \Comment{$\wbase$ apply on $\x^L$; $\wbase+\wlora$ on $\z$}
\State Sample \UnoDiff{ $\x^{L+1}\sim \mathbf{q}_0$} \Comment{free clean token from draft step}
\State Sample \UnoDiff{$\tilde{\x} \sim \mathbf{q}$} \Comment{$B-1$ tokens sampled in parallel}
\State $\mathbf{p}\gets \x_\wbase([\x^{L + 1}, \tilde{\x}]; \x)$
    \Comment{AR verifier distribution defined by $\wbase$}

\State sample
    $\mathbf{r} \overset{\mathrm{i.i.d.}}{\sim} \prod_{1}^{B-1}\mathcal{U}[0,1]$
\State $n\gets
    \min\!\left(
    \left\{
        i \in [B-1]:  r_i>
        \min\!\left(
            1,\frac{\mathbf{p}_i}{\UnoDiff{\mathbf{q}_i}}
        \right)
    \right\}
    \cup\{B\}
    \right)$

\If{$n=B$}
    \State sample $\tilde{\z} \sim \mathbf{p}_B$
\Else
    \State sample
        $\tilde{\z} \sim
        \operatorname{Norm}\!\left(
            [\mathbf{p}_n-\UnoDiff{\mathbf{q}_n}]_+
        \right)$
    
\EndIf
\State \textbf{return} $[\x, \x^{L+1}, \tilde{\x}^{1:n-1}, \tilde{\z}]$
\end{algorithmic}
\end{algorithm}

\section{Additional Experiments}
\subsection{Experiment Configs}
\subsubsection{Benchmarks}
We provide additional details about the set of benchmarks we evaluate \method{} on in \tab{tab:benchmark-overview}.

\begin{table*}[h]
\centering
\small
\caption{
Benchmarks used in our \method{} evaluation. ``Generations'' denotes the number of
responses sampled per example when computing average pass@1.
}
\label{tab:benchmark-overview}
\resizebox{\textwidth}{!}{%
\begin{tabular}{@{}lcp{0.62\textwidth}@{}}
\toprule
\textbf{Benchmark} & \textbf{Generations} & \textbf{Description} \\
\midrule

\multicolumn{3}{@{}l}{\textit{Agentic}} \\
\quad $\tau^2$-Bench
& 3
& Multi-turn tool-use tasks in which an agent interacts with a simulated user and domain APIs while following domain-specific policies. \\

\quad Terminal-Bench v2.1
& 3
& Realistic and complex tasks that autonomous agents complete in command-line container environments. \\

\quad SWE-bench Verified
& 3
& Human-validated software-engineering tasks requiring agents to resolve real GitHub issues by modifying their associated repositories. \\

\midrule
\multicolumn{3}{@{}l}{\textit{Long-Context Reasoning}} \\
\quad AA-LCR
& 3
& Open-answer questions requiring information extraction, synthesis, and reasoning over long documents such as reports and legal texts. \\

\midrule
\multicolumn{3}{@{}l}{\textit{Science and Knowledge}} \\
\quad AA-Omniscience
& 1
& Closed-book questions measuring factual recall and calibration across a broad range of economically relevant domains. \\

\quad Humanity's Last Exam
& 1
& Difficult expert-level questions spanning mathematics, the sciences, the humanities, and other academic disciplines. \\

\quad GPQA-Diamond
& 5
& Expert-validated, graduate-level multiple-choice questions in biology, physics, and chemistry. \\

\midrule
\multicolumn{3}{@{}l}{\textit{Math}} \\
\quad GSM8K
& 2
& Grade-school mathematics word problems requiring multi-step arithmetic reasoning. \\

\quad MATH500
& 2
& A 500-problem subset of MATH covering competition-level mathematical reasoning with free-form answers. \\

\quad AIME 2024--2026
& 10
& Integer-answer problems from the 2024, 2025, and 2026 American Invitational Mathematics Examinations. \\

\midrule
\multicolumn{3}{@{}l}{\textit{Coding}} \\
\quad MBPP
& 2
& Short Python program-synthesis problems specified through natural-language descriptions and test cases. \\
\quad HumanEval
& 2
& Hand-written Python function-completion problems evaluated using executable unit tests. \\

\midrule
\multicolumn{3}{@{}l}{\textit{Instruction Following}} \\
\quad IFEval
& 2
& Verifiable instruction-following tasks that test compliance with explicit, objectively checkable constraints. \\

\bottomrule
\end{tabular}
}
\end{table*}

\subsubsection{\methodqwen{}, \eagle{}, \dflash{} Sampler Configs for Qwen3-8B}\label{subsec:qwen-sampler-configs}
At $\temp=1$, we use top-$p=0.95$ and top-$k=50$; $\temp=0$ denotes greedy decoding. \methodqwen{} uses linear blocks of size $B\in\{4,8,16\}$ or, for tree verification, $B=16$, tree size $V=60$, and candidate top-$K=32$. DFlash uses blocks $B\in\{4,8,16\}$ and we evaluate both thinking-enabled and thinking-disabled variants (see~\supp{supp:dflash-thinking}). Linear EAGLE-3 $B\in\{4,8,16\}$ draft tokens; its tree configuration uses seven draft steps, top-$k=10$, and at most $V=60$ draft tokens.

\begin{table}[t]
    \centering
    \caption{Open-source model checkpoints used in our evaluations.}
    \label{tab:checkpoints}
    {\small    
    \begin{tabular}{l|l}
        \toprule
        Model &  Checkpoint \\
        \midrule
        \dg{} & \url{https://huggingface.co/google/diffusiongemma-26B-A4B-it} \\
        \nld{} & \url{https://huggingface.co/nvidia/Nemotron-Labs-Diffusion-14B} \\
        \eagle{} & \url{https://huggingface.co/AngelSlim/Qwen3-8B_eagle3} \\
        \dflash{} & \url{https://huggingface.co/z-lab/Qwen3-8B-DFlash-b16} \\
        Fast-dLLM v2 & \url{https://huggingface.co/Efficient-Large-Model/Fast_dLLM_v2_7B} \\
        SDAR & \url{https://huggingface.co/JetLM/SDAR-8B-Chat-b16} \\
        \bottomrule
    \end{tabular}
    }
\end{table}

\subsection{\method{} RL}
\paragraph{} Starting from the supervised fine-tuning (SFT) checkpoint, four specialized experts were trained for mathematics, code generation, tool use, and browse search using the DAPO reinforcement learning algorithm~\citep{yu2026dapo}. The Math expert was trained for 2,560 steps and generated 28.9B response tokens, while the Code expert was trained in two stages of 600 and 1,500 steps, producing 6B and 11.94B response tokens, respectively. Both experts used a learning rate of \(5 \times 10^{-7}\), 16 groups, and 32 prompts per step, yielding 512 sequences per step across 32 nodes per expert. Math training used maximum context lengths of 32,768 and 65,536 tokens, while Code training supported contexts up to 65,536 and 128K tokens. The Tool-use expert completed 39 steps with 128 prompts per step and generated 1.4B tokens, including tool-call tokens. The Browse Search expert completed 59 steps with 64 prompts per step and generated 8.4B tokens, including tool-call tokens. For both Tool-use and Browse Search, the maximum trajectory length was 128K tokens, covering model-generated tokens as well as tool outputs. Finally, the four experts were consolidated into a single model using ISO-Merger (RAM; \citeauthor{yuan2026behavior}, \citeyear{yuan2026behavior}), a data-free approach that combines specialists trained from a shared base checkpoint without requiring additional rollouts, gradient updates, or distillation.

\subsection{\method{} Evaluations}

\begin{table}[H]
\caption{We report TPFs for a subset of the evaluations using different sampler configurations for the \method{} model. Among all configurations, the Linear sampler with $B=4$ achieves the highest system throughput at a batch size of 64, while the Tree sampler with $(B,K,V)=(16,32,32)$ achieves the highest per-user throughput (batch size 1).}
\label{tab:uno-sampler-ablations}
\centering
\small
\begin{tabular}{lccc|ccc}
\toprule
Dataset
& \shortstack{Linear \\B4}
& \shortstack{Linear \\B8}
& \shortstack{Linear \\B16}
& \shortstack{Tree \\B16,K32,V32}
& \shortstack{Tree \\B16,K64,V64}
& \shortstack{Tree \\B16,K64,V32} \\
\midrule
GSM8K
& 1.9 & 2.3 & 2.4
& 2.7 & 2.6 & 2.8 \\

AIME-24
& 1.8 & 2.1 & 2.2
& 2.5 & 2.5 & 2.4 \\

AIME-25
& 1.8 & 2.2 & 2.3
& 2.4 & 2.6 & 2.6 \\

AIME-26
& 1.8 & 2.1 & 2.1
& 2.4 & 2.5 & 2.4 \\

MATH500
& 1.8 & 2.1 & 2.2
& 2.4 & 2.5 & 2.5 \\

HumanEval
& 1.8 & 2.6 & 2.8
& 3.4 & 2.5 & 3.4 \\

AA-LCR
& 1.8 & 2.1 & 2.2
& 2.6 & 2.5 & 2.4 \\
\midrule

TPF
& 1.8 & 2.2 & 2.3
& 2.6 & 2.5 & 2.6 \\
Highest Sys. Throughput
& \textbf{5190}  &    5034  &    3445 &      2665 &     1573 &      2709 \\
Highest Per-req. Throughput
& 275 &      319&    345&        \textbf{379}&      339&       371\\
\bottomrule
\end{tabular}
\end{table}

\begin{table}[H]
    \caption{DG, Uno, NLD, Mercury2. $^*$As reported by Artificial Analysis' live tracker on August 30, 2026.}
    \label{tab:uno-baseline-throughput}
    \centering
    {\small
    \begin{tabular}{c|cc}
        \toprule
        Method & Max. Sys. Throughput (toks / sec) & Max. Per-req. Throughput (toks / sec) \\
        \midrule
        \method{} (Ours)  & 5255 & 383 \\
        AR (Ours)  & 3577 & 176\\
        \dg{}  & 1136 & 836\\
        \nld{} & 2794 & 290 \\
        Mercury-2 & 1197$^*$ & 769$^*$ \\
        \bottomrule
    \end{tabular}
    }
\end{table}

\begin{table}[H]
\caption{TPFs for the SFT checkpoint and its RL-post-trained counterpart, evaluated using the diffusion weights trained for the SFT checkpoint. \textbf{Even after extensive RL post-training, the SFT diffusion adapters retained their speedup}, with only a $6\%$ reduction in TPFs.}
\label{tab:sft-rl-tpf-results}
\centering
\small
\setlength{\tabcolsep}{5pt}
\begin{tabular}{lcc}
\toprule
& {SFT}
& {Post-Trained} \\
\midrule

\multicolumn{3}{l}{\textit{Math}} \\
\quad GSM8K   & $2.66$ & $1.95$ \\
\quad MATH500 & $2.27$ & $1.93$ \\
\quad AIME-24 & $2.17$ & $1.93$ \\
\quad AIME-25 & $2.26$ & $1.97$ \\
\quad AIME-26 & $2.22$ & $1.83$ \\
\midrule

\multicolumn{3}{l}{\textit{Coding}} \\
\quad HumanEval & $1.97$ & $2.34$ \\
\quad MBPP      & $2.35$ & $2.19$ \\
\midrule

\multicolumn{3}{l}{\textit{Science and Knowledge}} \\
\quad GPQA-Diamond & $2.03$ & $1.97$ \\
\midrule

\multicolumn{3}{l}{\textit{Instruction Following}} \\
\quad IFEval & $2.12$ & $2.50$ \\
\midrule

\multicolumn{3}{l}{\textit{Other Benchmarks}} \\
\quad HLE            & $2.14$ & $2.15$ \\
\quad AA-Omniscience & $2.57$ & $2.34$ \\
\midrule

{TPF}
& ${2.25}$
& ${2.10}$ \\
\bottomrule
\end{tabular}

\end{table}

\subsection{Qwen Finetuned on OpenThoughts}
\tab{tab:qwen-openthoughs-ablation} compares Qwen3-8B with a variant fine-tuned on OpenThoughts across a subset of benchmarks. Fine-tuning on OpenThoughts substantially degrades performance, reducing accuracy by up to $15\%$ points, depending on the benchmark. Nevertheless, diffusion weights trained on OpenThoughts successfully accelerate generation despite the distribution mismatch between OpenThoughts and the data used to train the AR weights.

\begin{table}[H]
\centering
\caption{This table compares Qwen3-8B with a variant fine-tuned on OpenThoughts across a subset of benchmarks.}
\label{tab:qwen-openthoughs-ablation}
\begin{tabular}{lcc}
\toprule
& Qwen3-8B & Qwen-finetuned \\
\midrule
\textit{Math} & & \\
\quad GSM8K      & $96_{\pm 0.2}$  & $95.2_{\pm 1.1}$ \\
\quad MATH500    & $96.2_{\pm 0.3}$  & $94.8_{\pm 0.0}$ \\
\quad AIME-24    & $77.7_{\pm 3.0}$  & $68.3_{\pm 17.6}$ \\
\quad AIME-25    & $70.7_{\pm 4.2}$  & $53.3_{\pm 0.0}$ \\
\quad AIME-26    & $67.7_{\pm 3.9}$ & $61.7_{\pm 5.9}$ \\
\midrule
\textit{Coding} & & \\
\quad HumanEval  & $94.4_{\pm 1.0}$  & $77.8_{\pm 4.6}$ \\
\quad MBPP       & $88.7_{\pm 0.4}$  & $78.1_{\pm 0.4}$ \\
\quad LCBv6      & $50.9_{\pm 1.2}$  & $42.5_{\pm 0.8}$ \\
\bottomrule
\end{tabular}
\end{table}

\subsection{\texorpdfstring{\methodqwen{}}{Uno-Qwen} vs I-DLM (R-ISD) Comparison}\label{supp:I-DLM}
In this section, we compare I-DLM (R-ISD), the lossless variant of I-DLM, with our method, \methodqwen{}.

In the first experiment, we train I-DLM (R-ISD) with a rank-128 LoRA adapter for one epoch on the OpenThoughts dataset, using exactly the same training configuration as \methodqwenone{}. For inference, we use the sampler provided in the official I-DLM repository\footnote{\url{https://github.com/Introspective-Diffusion/I-DLM/commit/a23c1a12ef997c7f3ad616b25bcfb62db39ded68}}. All results for I-DLM (R-ISD) in \tab{tab:original-I-DLM-sampler-comparison} are evaluated with a block size of $B=16$. The results show that the I-DLM sampler degrades accuracy on numerous benchmarks relative to the Qwen AR model, as highlighted in red, whereas \methodqwenone{} does not. Upon inspecting the I-DLM codebase, we find that its sampler performs greedy drafting\footnote{\url{https://github.com/Introspective-Diffusion/I-DLM/blob/a23c1a12ef997c7f3ad616b25bcfb62db39ded68/inference/sglang/sglang/srt/dllm/algorithm/idlm_blockN.py\#L575}}, but its rejection-sampling verification procedure is not adjusted accordingly. \textbf{As a result, the sampler does not preserve the claimed losslessness property}.

I-DLM (R-ISD) uses LoRA adapters and retains the transformer's causal attention pattern. It is therefore compatible with our $\Psi$-Spec sampler using the masked prior $\prior = \m$. In the second set of experiments, we evaluate I-DLM (R-ISD) with $\Psi$-Spec. As shown in \tab{tab:original-I-DLM-sampler-comparison}, this combination remains lossless. We observe the same result when evaluating the I-DLM (R-ISD) adapters released in the official repository. However, both variants provide considerably lower speedups than our method.
\begin{table}[H]
    \centering
    \caption{
I-DLM and \methodqwenone{} evaluation with the linear sampler at $B=16$ and $\temp=1$. We compare the I-DLM checkpoints (Ours and Official) under $\Psi$-spec samplers with $\prior = \m$; accuracy is reported
with 95\% confidence intervals across seeds, using two seeds for MMLU-Pro and
ten seeds for all other benchmarks.
}
\label{tab:original-I-DLM-sampler-comparison}
\resizebox{\textwidth}{!}{%
\begin{tabular}{lccccccccc}
\toprule
\textit{Sampler} 
& 
& \multicolumn{2}{c}{I-DLM sampler}
& \multicolumn{6}{c}{$\boldsymbol{\Psi}$-spec sampler} \\
\cmidrule(lr){3-4}\cmidrule(lr){5-10}
\textit{Model}
& \multicolumn{1}{c}{Qwen3 AR}
& \multicolumn{2}{c}{I-DLM (R-ISD) (ours)}
& \multicolumn{2}{c}{I-DLM (R-ISD) (ours)}
& \multicolumn{2}{c}{I-DLM (R-ISD) (official)}
& \multicolumn{2}{c}{\method{} (ours)} \\
\cmidrule(lr){3-4}\cmidrule(lr){5-6}\cmidrule(lr){7-8}\cmidrule(lr){9-10}
& Acc. (\%) & Acc. (\%) & TPF & Acc. (\%) & TPF & Acc. (\%) & TPF & Acc. (\%) & TPF \\
\midrule
\textit{Math} & & & & & & & & & \\
\quad GSM8K & $96.0_{\pm 0.2}$ & $96.0_{\pm 0.2}$ & 2.14 & $96.0_{\pm 0.2}$ & 1.95 & $96.0_{\pm 0.2}$ & 2.23 & $96.0_{\pm 0.2}$ & 2.66 \\
\quad MATH500 & $96.2_{\pm 0.3}$ & $96.3_{\pm 0.5}$ & 2.36 & $96.1_{\pm 0.4}$ & 2.06 & $95.9_{\pm 0.3}$ & 2.06 & $96.2_{\pm 0.3}$ & 2.97 \\
\quad AIME-24 & $77.7_{\pm 3.0}$ & $77.3_{\pm 2.9}$ & 2.33 & $76.7_{\pm 4.2}$ & 2.03 & $76.7_{\pm 2.8}$ & 1.80 & $77.7_{\pm 2.8}$ & 2.85 \\
\quad AIME-25 & $70.7_{\pm 4.2}$ & \newnewourredcell{$63.0_{\pm 3.6}$} & 2.38 & $68.7_{\pm 4.8}$ & 2.06 & $70.7_{\pm 5.1}$ & 1.79 & $69.7_{\pm 3.2}$ & 2.92 \\
\quad AIME-26 & $67.7_{\pm 3.9}$ & $68.7_{\pm 4.7}$ & 2.34 & $64.3_{\pm 3.2}$ & 2.02 & $66.7_{\pm 3.2}$ & 1.78 & $67.0_{\pm 3.3}$ & 2.84 \\
\midrule
\textit{Coding} & & & & & & & & & \\
\quad HumanEval & $94.4_{\pm 1.0}$ & $93.7_{\pm 0.9}$ & 2.30 & $95.0_{\pm 0.6}$ & 1.77 & $94.5_{\pm 0.5}$ & 1.98 & $94.3_{\pm 0.7}$ & 2.40 \\
\quad MBPP & $88.7_{\pm 0.4}$ & \newnewourredcell{$87.8_{\pm 0.4}$} & 2.19 & {$88.7_{\pm 0.7}$} & 1.80 & $88.7_{\pm 0.5}$ & 1.93 & $88.8_{\pm 0.4}$ & 2.47 \\
\quad LCBv6 & $50.9_{\pm 1.2}$ & \newnewourredcell{$48.9_{\pm 1.5}$} & 2.05 & $50.4_{\pm 1.4}$ & 1.68 & $51.0_{\pm 1.4}$ & 1.60 & $50.1_{\pm 1.4}$ & 2.30 \\
\midrule
\textit{Science and Knowledge} & & & & & & & & & \\
\quad GPQA & $56.8_{\pm 0.7}$ & $57.2_{\pm 1.1}$ & 2.01 & $56.1_{\pm 1.1}$ & 1.66 & $56.6_{\pm 1.1}$ & 1.71 & $57.0_{\pm 1.0}$ & 2.35 \\
\quad GPQA-Diamond & $59.5_{\pm 0.9}$ & \newnewourredcell{$58.4_{\pm 0.5}$} & 2.02 & $59.4_{\pm 1.4}$ & 1.67 & $60.6_{\pm 1.2}$ & 1.69 & $59.7_{\pm 1.1}$ & 2.35 \\
\quad MMLU-Pro & $74.7_{\pm 0.7}$ & $74.6_{\pm 0.5}$ & 2.07 & $74.7_{\pm 0.2}$ & 1.75 & $75.1_{\pm 0.9}$ & 1.84 & $74.7_{\pm 0.7}$ & 2.46 \\
\midrule
\textit{Instruction Following} & & & & & & & & & \\
\quad IFEval & $85.8_{\pm 0.7}$ & $85.3_{\pm 0.7}$ & 1.87 & $85.9_{\pm 0.8}$ & 1.65 & $85.7_{\pm 0.6}$ & 1.85 & $86.0_{\pm 0.7}$ & 1.96 \\
\midrule
Average & $76.6_{\pm 0.4}$ & $75.6_{\pm 0.6}$ & 2.17 & $76.0_{\pm 0.7}$ & 1.84 & $76.5_{\pm 0.5}$ & 1.86 & $76.4_{\pm 0.4}$ & 2.55 \\
\bottomrule
\end{tabular}%
}
\end{table}

Additionally, \tab{tab:official-I-DLM-psi-tpf-by-block} provides TPF for the I-DLM LoRA checkpoint evaluated on block sizes $B=4$, $B=8$, and $B=16$. The results show a significant increase in TPF from $B=4$ to $B=8$, while TPF seems to saturate beyond $B=8$.

\begin{table}[H]
    \centering
    \caption{TPFs for the official I-DLM (R-ISD) model with our $\Psi$-spec samplers across block sizes $B$.}
    \label{tab:official-I-DLM-psi-tpf-by-block}
    \begin{tabular}{lccc}
    \toprule
    & $B$=$4$ & $B$=$8$ & $B$=$16$ \\
    \midrule
    \textit{Math} & & & \\
    \quad GSM8K & 1.87 & 2.20 & 2.23 \\
    \quad MATH500 & 1.80 & 2.04 & 2.06 \\
    \quad AIME-24 & 1.68 & 1.80 & 1.80 \\
    \quad AIME-25 & 1.66 & 1.81 & 1.79 \\
    \quad AIME-26 & 1.67 & 1.78 & 1.78 \\
    \midrule
    \textit{Coding} & & & \\
    \quad HumanEval & 1.75 & 1.96 & 1.98 \\
    \quad MBPP & 1.73 & 1.93 & 1.93 \\
    \quad LCBv6 & 1.54 & 1.60 & 1.60 \\
    \midrule
    \textit{Science and Knowledge} & & & \\
    \quad GPQA & 1.61 & 1.71 & 1.71 \\
    \quad GPQA-Diamond & 1.59 & 1.69 & 1.69 \\
    \quad MMLU-Pro & 1.69 & 1.83 & 1.84 \\
    \midrule
    \textit{Instruction Following} & & & \\
    \quad IFEval & 1.73 & 1.86 & 1.85 \\
    \midrule
    Average & 1.69 & 1.85 & 1.86 \\
    \bottomrule
    \end{tabular}
\end{table}

\subsection{\texorpdfstring{\methodqwen{}}{Uno-Qwen} Ablations}

\subsubsection{LoRA Rank and Loss Ablations}\label{supp:lora-ablation}

In \tab{tab:qwen-owt-ablation}, we ablate the LoRA rank of \methodqwenone{}, as well as various weightings of KL and TV loss terms. We see that average TPF increases from 2.39 to 2.47 when increasing the LoRA rank from 128 to 256, while the number of trainable parameters increases from 349M to 698M. It turns out that a loss weighting of $0.01\times$KL plus $1\times$TV slightly outperforms a pure TV loss objective. Note that a factor of $0.01$ for the KL term amounts to a KL loss term only about $10\times$ smaller in magnitude than the TV term, as the KL loss is naturally around $10\times$ larger than the TV loss.

\begin{table*}[h]
\centering
\caption{
TPF results for one-epoch \methodqwenone{} checkpoints trained with TV-only, KL-only, and weighted KL+TV objectives. The TV-only objective is evaluated using rank-128 LoRA adapters (the default \methodqwen{} configuration, highlighted in blue) and rank-256 adapters, while all other objectives use rank-128 adapters. All adapters are applied to every projection and use $\alpha_{\mathrm{LoRA}}/r=2$. All checkpoints are evaluated using the linear sampler with $B=16$ and $\temp=1$.
}
\label{tab:qwen-owt-ablation}
\resizebox{\textwidth}{!}{%
\begin{tabular}{lZcccccc}
\toprule
& \multicolumn{2}{c}{TV only}
& \multicolumn{5}{c}{Loss ablations ($r$=$128$)} \\
\cmidrule(lr){2-3}
\cmidrule(lr){4-8}
& \shortstack{$r$=$128$}
& \shortstack{$r$=$256$}
& KL only
& $\mathrm{KL}+\mathrm{TV}$
& $0.1\,\mathrm{KL}+\mathrm{TV}$
& $0.05\,\mathrm{KL}+\mathrm{TV}$
& $0.01\,\mathrm{KL}+\mathrm{TV}$ \\
& \multicolumn{2}{c}{$\alpha$=0, $\beta$=1}
& $\alpha$=1, $\beta$=0
& $\alpha$=1, $\beta$=1
& $\alpha$=0.1, $\beta$=1
& $\alpha$=0.05, $\beta$=1
& $\alpha$=0.01, $\beta$=1 \\
\midrule
\textit{Math} & & & & & & & \\
\quad GSM8K
& 2.47 & 2.53 & 2.27 & 2.30 & 2.39 & 2.42 & 2.46 \\
\quad MATH500
& 2.73 & 2.82 & 2.52 & 2.56 & 2.65 & 2.70 & 2.71 \\
\quad AIME-24
& 2.65 & 2.75 & 2.47 & 2.51 & 2.62 & 2.66 & 2.70 \\
\quad AIME-25
& 2.76 & 2.80 & 2.55 & 2.56 & 2.66 & 2.72 & 2.75 \\
\quad AIME-26
& 2.81 & 2.82 & 2.53 & 2.51 & 2.59 & 2.67 & 2.68 \\
\midrule
\textit{Coding} & & & & & & & \\
\quad HumanEval
& 2.26 & 2.37 & 2.04 & 2.07 & 2.30 & 2.23 & 2.25 \\
\quad MBPP
& 2.33 & 2.40 & 2.13 & 2.14 & 2.26 & 2.29 & 2.32 \\
\quad LCBv6
& 2.17 & 2.22 & 1.94 & 1.97 & 2.08 & 2.13 & 2.17 \\
\midrule
\textit{Science and Knowledge} & & & & & & & \\
\quad GPQA
& 2.21 & 2.27 & 1.99 & 2.03 & 2.15 & 2.19 & 2.20 \\
\quad GPQA-Diamond
& 2.22 & 2.26 & 2.00 & 2.03 & 2.15 & 2.18 & 2.21 \\
\quad MMLU-Pro
& 2.31 & 2.36 & 2.11 & 2.14 & 2.24 & 2.28 & 2.30 \\
\midrule
\textit{Instruction Following} & & & & & & & \\
\quad IFEval
& 1.79 & 1.99 & 2.08 & 1.93 & 2.08 & 2.08 & 2.02 \\
\midrule
Average TPF
& 2.39 & 2.47 & 2.22 & 2.23 & 2.35 & 2.38 & 2.40 \\
\bottomrule
\end{tabular}
}
\end{table*}

\subsubsection{Training Block Size Curriculum}\label{supp:block-curricula}
Here we ablate two block-size curricula for \methodqwen{} training. \tab{tab:block-curricula} shows the TPF attained by two different three-epoch block-size curricula, both initialized from a model trained on block size 2 for 0.5 epochs, and block size 4 for another 0.5 epochs (written as $0.5@2,\ 0.5@4$). The first curriculum is our default curriculum used for \methodqwen{}, which continues at $0.5@6,\ 0.5@8$, $0.5@12$, and $0.5@16$; the second one continues at $2@16$. Average TPF drops from 2.71 to 2.65 when employing the second curriculum, which shows that incrementally increasing block size during training is beneficial over consistently training at a large block size.

\begin{table*}[h]
\centering
\caption{
TPF for two three-epoch TV-only block curricula. The notation $e@B$
denotes $e$ epochs of training at block size $B$. Both models first train on
$0.5@2,\ 0.5@4$. The standard curriculum then progressively increases the
block size, whereas the fixed-$B=16$ curriculum trains for the remaining two
epochs entirely at $B=16$. Models are evaluated with the linear sampler at
$B=16$ and $\temp=1$.
}
\label{tab:block-curricula}
\begin{tabular}{lcc}
\toprule
& \shortstack{
$0.5@2,\ 0.5@4,$\\
$0.5@6,\ 0.5@8,\ 0.5@12,\ 0.5@16$}
& \shortstack{
$0.5@2,\ 0.5@4,$\\
$2@16$} \\
\midrule
\textit{Math} & & \\
\quad GSM8K        & 2.83 & 2.81 \\
\quad MATH500      & 3.13 & 3.10 \\
\quad AIME-24      & 3.18 & 3.04 \\
\quad AIME-25      & 3.14 & 3.00 \\
\quad AIME-26      & 3.03 & 3.02 \\
\midrule
\textit{Coding} & & \\
\quad HumanEval    & 2.58 & 2.49 \\
\quad MBPP         & 2.60 & 2.58 \\
\quad LCBv6        & 2.37 & 2.37 \\
\midrule
\textit{Science and Knowledge} & & \\
\quad GPQA         & 2.45 & 2.45 \\
\quad GPQA-Diamond & 2.47 & 2.46 \\
\quad MMLU-Pro     & 2.57 & 2.56 \\
\midrule
\textit{Instruction Following} & & \\
\quad IFEval       & 2.20 & 1.95 \\
\midrule
Average             & 2.71 & 2.65 \\
\bottomrule
\end{tabular}
\end{table*}

\clearpage
\subsubsection{LoRA Adapter Ablations}\label{supp:lora-adapter}
We ablate which projection matrices in the transformer layer to apply LoRA weights to in \tab{tab:lora-target-matrices}. By default, we apply LoRA to all projections (this includes the Q, K, V, O matrices from the attention layer, as well as the MLP gate, up, and down projections); in addition, we ablate only applying LoRA to the attention layers (i.e., Q, K, V, and O projections); only to the Q and K projections of the attention layer; only to the Q projections; and only to the O projections, which is the setup recommended in \citet{fu2026nemotronlabsdiffusion}. In each setting, we adapted the rank to keep parameter parity at 349M LoRA parameters. All models for this ablation were trained on one epoch at block size eight.
\tab{tab:lora-target-matrices} shows that at fixed $\alpha_\text{LoRA}\ / \ \text{rank}=2$, applying LoRA to all projections work best. Consequently, we apply LoRA all to projections for our \method{} models.

\begin{table*}[h]
\centering
\caption{
TPF for \methodqwenone{} LoRA target-projection ablations
with $\alpha_{\mathrm{LoRA}}/r_{\mathrm{LoRA}}=2$.
Ranks are chosen to approximately maintain LoRA parameter parity.
All checkpoints are evaluated with the linear sampler at $B$=$16$ and
$\temp=1$.
}
\label{tab:lora-target-matrices}
\resizebox{\textwidth}{!}{%
\begin{tabular}{lccccc}
\toprule
& \shortstack{All\\projections}
& \shortstack{$Q,K,V,O$\\projections}
& \shortstack{$Q,V$\\projections}
& $Q$ projection
& $O$ projection \\
\midrule
\textit{Rank $r$} & 128 & 364 & 729 & 1184 & 1184 \\
\textit{LoRA $\alpha_{\mathrm{LoRA}}$} & 256 & 728 & 1458 & 2368 & 2368 \\
\textit{$\alpha_{\mathrm{LoRA}}/r$} & 2 & 2 & 2 & 2 & 2 \\
\midrule
\textit{Math} & & & & & \\
\quad GSM8K        & 2.47 & 2.45 & 2.32 & 2.20 & 2.43 \\
\quad MATH500      & 2.73 & 2.68 & 2.55 & 2.40 & 2.69 \\
\quad AIME-24      & 2.65 & 2.67 & 2.48 & 2.35 & 2.72 \\
\quad AIME-25      & 2.76 & 2.70 & 2.54 & 2.39 & 2.71 \\
\quad AIME-26      & 2.81 & 2.64 & 2.46 & 2.32 & 2.68 \\
\midrule
\textit{Coding} & & & & & \\
\quad HumanEval    & 2.26 & 2.23 & 2.16 & 2.11 & 2.33 \\
\quad MBPP         & 2.33 & 2.31 & 2.22 & 2.11 & 2.33 \\
\quad LCBv6        & 2.17 & 2.15 & 2.05 & 1.95 & 2.15 \\
\midrule
\textit{Science and Knowledge} & & & & & \\
\quad GPQA         & 2.21 & 2.19 & 2.10 & 2.02 & 2.20 \\
\quad GPQA-Diamond & 2.22 & 2.21 & 2.11 & 2.01 & 2.21 \\
\quad MMLU-Pro     & 2.31 & 2.29 & 2.20 & 2.09 & 2.30 \\
\midrule
\textit{Instruction Following} & & & & & \\
\quad IFEval       & 1.79 & 2.00 & 1.88 & 1.73 & 1.78 \\
\midrule
\textbf{Average}  & 2.39 & 2.38 & 2.26 & 2.14 & 2.38 \\
\bottomrule
\end{tabular}%
}
\end{table*}

\subsubsection{Varying LoRA $\alpha / r$}\label{supp:lora-alpha}
For LoRA with rank $r_{\text{LoRA}}$ and a projection $W_\text{base}\in\mathbb{R}^{n\times m}$ of the base model, both LoRA training and inference use
\begin{equation}
    W = W_{\text{base}}+\frac{\alpha_{\text{LoRA}}}{r_{\text{LoRA}}}BA,
\end{equation}
where $B\in\mathbb{R}^{n\times r}$, $A\in\mathbb{R}^{r\times m}$, and $\alpha_\text{LoRA}>0$ is a hyperparameter that controls the overall contribution of the LoRA adapter. In \tab{tab:lora-alpha-comparison} we ablate the ratio $\alpha_\text{LoRA}/r_{\text{LoRA}}$ across values between 2 and 256 or both one-epoch and three-epoch models. 
All three-epoch models in this ablation were trained on the block size curriculum described in \supp{supp:block-curricula}.

The table shows that the optimal ratio differs depending on the training horizon, and $\alpha_{\text{LoRA}}/r_{\text{LoRA}}=16$ is optimal at three epochs, while $\alpha_{\text{LoRA}}/r_{\text{LoRA}}=64$ is optimal at one epoch. Consequently, we pick $\alpha_{\text{LoRA}}/r_{\text{LoRA}}=16$ for \methodqwen{}, while for \method{}, we set $\alpha_{\text{LoRA}}/r_{\text{LoRA}}=64$ which we observed to be preferable.

\begin{table*}[p]
\centering
\caption{
TPF for one-epoch and three-epoch, TV-only \methodqwenone{} models with
rank-128 LoRA adapters applied to all projections, across LoRA scaling
values. All checkpoints are evaluated with the linear sampler at $B=16$
and $\temp=1$; averages are unweighted across the 12 benchmarks.
}
\label{tab:lora-alpha-comparison}

\begin{subtable}[t]{\textwidth}
\centering
\caption{One epoch of training.}
\label{tab:lora-alpha-comparison-one-epoch}
\resizebox{\textwidth}{!}{%
\begin{tabular}{lccccccc}
\toprule
& $\alpha_{\mathrm{LoRA}}=256$
& $\alpha_{\mathrm{LoRA}}=512$
& $\alpha_{\mathrm{LoRA}}=1024$
& $\alpha_{\mathrm{LoRA}}=2048$
& $\alpha_{\mathrm{LoRA}}=8192$
& $\alpha_{\mathrm{LoRA}}=16384$
& $\alpha_{\mathrm{LoRA}}=32768$ \\
\midrule
\textit{$\alpha_{\mathrm{LoRA}}/r$}
& 2 & 4 & 8 & 16 & 64 & 128 & 256 \\
\midrule
\textit{Math} & & & & & & & \\
\quad GSM8K
& 2.47 & 2.53 & 2.63 & 2.66 & 2.76 & 2.76 & 2.72 \\
\quad MATH500
& 2.73 & 2.79 & 2.89 & 2.97 & 3.04 & 3.02 & 2.98 \\
\quad AIME-24
& 2.65 & 2.77 & 2.85 & 2.85 & 3.00 & 2.92 & 2.90 \\
\quad AIME-25
& 2.76 & 2.83 & 2.91 & 2.92 & 3.02 & 3.01 & 2.90 \\
\quad AIME-26
& 2.81 & 2.72 & 2.77 & 2.84 & 2.93 & 2.93 & 2.82 \\
\midrule
\textit{Coding} & & & & & & & \\
\quad HumanEval
& 2.26 & 2.39 & 2.36 & 2.40 & 2.55 & 2.45 & 2.51 \\
\quad MBPP
& 2.33 & 2.39 & 2.43 & 2.47 & 2.56 & 2.51 & 2.46 \\
\quad LCBv6
& 2.17 & 2.21 & 2.26 & 2.30 & 2.32 & 2.31 & 2.25 \\
\midrule
\textit{Science and Knowledge} & & & & & & & \\
\quad GPQA
& 2.21 & 2.26 & 2.31 & 2.35 & 2.41 & 2.41 & 2.37 \\
\quad GPQA-Diamond
& 2.22 & 2.27 & 2.32 & 2.35 & 2.40 & 2.40 & 2.36 \\
\quad MMLU-Pro
& 2.31 & 2.36 & 2.41 & 2.46 & 2.51 & 2.50 & 2.45 \\
\midrule
\textit{Instruction Following} & & & & & & & \\
\quad IFEval
& 1.79 & 2.01 & 2.13 & 1.96 & 2.01 & 2.27 & 1.84 \\
\midrule
\textbf{Average}
& 2.39 & 2.46 & 2.52 & 2.55 & 2.63 & 2.62 & 2.54 \\
\bottomrule
\end{tabular}%
}
\end{subtable}

\vspace{1em}

\begin{subtable}[t]{\textwidth}
\centering
\caption{Three epochs of training.}
\label{tab:lora-alpha-comparison-three-epochs}
\begin{tabular}{lccc}
\toprule
& $\alpha_{\mathrm{LoRA}}=2048$
& $\alpha_{\mathrm{LoRA}}=8192$
& $\alpha_{\mathrm{LoRA}}=16384$ \\
\midrule
\textit{$\alpha_{\mathrm{LoRA}}/r$} & 16 & 64 & 128 \\
\midrule
\textit{Math} & & & \\
\quad GSM8K        & 2.83 & 2.83 & 2.78 \\
\quad MATH500      & 3.13 & 3.12 & 3.06 \\
\quad AIME-24      & 3.18 & 3.02 & 2.94 \\
\quad AIME-25      & 3.14 & 3.08 & 3.00 \\
\quad AIME-26      & 3.03 & 3.05 & 2.97 \\
\midrule
\textit{Coding} & & & \\
\quad HumanEval    & 2.58 & 2.49 & 2.44 \\
\quad MBPP         & 2.60 & 2.56 & 2.51 \\
\quad LCBv6        & 2.37 & 2.34 & 2.30 \\
\midrule
\textit{Science and Knowledge} & & & \\
\quad GPQA         & 2.45 & 2.43 & 2.41 \\
\quad GPQA-Diamond & 2.47 & 2.47 & 2.39 \\
\quad MMLU-Pro     & 2.57 & 2.55 & 2.50 \\
\midrule
\textit{Instruction Following} & & & \\
\quad IFEval       & 2.20 & 2.00 & 2.13 \\
\midrule
\textbf{Average}  & 2.71 & 2.66 & 2.62 \\
\bottomrule
\end{tabular}
\end{subtable}

\end{table*}

\clearpage
\subsection{DFlash Thinking Mode Ablation}\label{supp:dflash-thinking}

In \tab{tab:dflash-thinking-ablation} we evaluate DFlash~\citep{chen2026dflash} with thinking mode enabled and disabled, at $\temp=1$. Our results show that while disabling thinking mode significantly increases tokens per step, it deteriorates accuracy from around $76\%$ to roughly $55\%$. This quality difference
does not contradict losslessness: enabling thinking changes the chat template
and hence the target-model distribution that the speculative decoder
preserves. In light of these results, we subsequently enable thinking mode for DFlash~\citep{zlab2026dflashmodelcard}.

\begin{table*}[h]
\centering
\caption{
\textbf{DFlash thinking ablation} at $\temp=1$, top-$p=0.95$, and top-$k=50$.
Accuracy in percent. AL denotes average acceptance length, i.e., tokens per verification step.
}
\label{tab:dflash-thinking-ablation}
\resizebox{\textwidth}{!}{%
\begin{tabular}{lcccccccc}
\toprule
\multirow{2}{*}{ }
& \multicolumn{4}{c}{{DFlash,} $B$=$4$}
& \multicolumn{4}{c}{{DFlash,} $B$=$16$}\\
\cmidrule(lr){2-5}
\cmidrule(lr){6-9}
& \multicolumn{2}{c}{{No thinking}}
& \multicolumn{2}{c}{{Thinking}}
& \multicolumn{2}{c}{{No thinking}}
& \multicolumn{2}{c}{{Thinking}} \\
\cmidrule(lr){2-3}
\cmidrule(lr){4-5}
\cmidrule(lr){6-7}
\cmidrule(lr){8-9}
& Acc. (\%) & AL
& Acc. (\%) & AL
& Acc. (\%) & AL
& Acc. (\%) & AL \\
\midrule
\textit{Math} & & & & & & & & \\
\quad GSM8K
& 93.93 & 3.15 & 95.91 & 2.47
& 93.56 & 5.75 & 95.83 & 3.38 \\

\quad MATH500
& 84.00 & 3.19 & 96.60 & 2.28
& 82.80 & 6.23 & 96.20 & 3.24 \\

\quad AIME-24
& 23.33 & 2.95 & 76.67 & 1.96
& 30.00 & 5.41 & 76.67 & 2.76 \\

\quad AIME-25
& 20.00 & 2.95 & 60.00 & 1.90
& 16.67 & 4.88 & 73.33 & 2.56 \\

\quad AIME-26
& 16.67 & 3.01 & 63.33 & 1.93
& 20.00 & 4.76 & 66.67 & 2.57 \\

\midrule
\textit{Coding} & & & & & & & & \\
\quad HumanEval
& 87.20 & 3.68 & 94.51 & 2.35
& 87.20 & 10.10 & 94.51 & 3.03 \\

\quad MBPP
& 67.40 & 3.10 & 89.60 & 2.22
& 67.40 & 5.17 & 90.00 & 3.04 \\

\quad LCBv6
& 25.14 & 2.64 & 49.71 & 1.76
& 24.00 & 4.90 & 49.71 & 2.22 \\

\midrule
\textit{Science and Knowledge} & & & & & & & & \\
\quad GPQA
& 43.53 & 2.57 & 57.59 & 1.98
& 44.42 & 3.61 & 55.80 & 2.57 \\

\quad GPQA-Diamond
& 43.43 & 2.50 & 61.11 & 1.92
& 44.44 & 3.70 & 58.08 & 2.50 \\

\quad MMLU-Pro
& 66.45 & 2.70 & 74.75 & 2.11
& 66.53 & 4.12 & 74.83 & 2.80 \\

\midrule
\textit{Instruction Following} & & & & & & & & \\
\quad IFEval
& 85.21 & 1.84 & 86.14 & 1.92
& 87.80 & 2.59 & 84.66 & 2.26 \\

\midrule
Average
& 54.69 & \textbf{2.86}
& \textbf{75.49} & 2.07
& 55.40 & \textbf{5.10}
& \textbf{76.36} & 2.74 \\
\bottomrule
\end{tabular}%
}
\end{table*}

\subsection{Extended Comparison vs EAGLE-3 and DFlash}\label{supp:eagle-dflash}
By default, we evaluate DFlash with block sizes $B \in \{8,16\}$. For EAGLE-3, we use a linear block size of $B=8$, as well as its standard tree configuration with depth $D=7$ (corresponding to block size $B=8$), top-$k=10$, and a verification budget of $V=60$ tokens. These EAGLE-3 tree parameters follow the official settings provided in the paper and repository.

We provide extended evaluation results for \methodqwen{}, EAGLE-3, and DFlash across various sampler settings in \tab{tab:acceptance-by-method-raw}.

\begin{table*}[p]
\centering
\caption{
Tokens per step across samplers for \methodqwen{}, EAGLE-3, and
thinking-enabled DFlash. Values at both $\temp=0$ and $\temp=1$ are
unfiltered.
}
\label{tab:acceptance-by-method-raw}

\begin{subtable}{\textwidth}
\centering
\caption{
\methodqwen{} linear and tree samplers.
}
\label{tab:uno-acceptance-by-block-unfiltered}
\resizebox{\textwidth}{!}{%
\begingroup
\renewcommand{\arraystretch}{0.85}
\begin{tabular}{l*{10}{c}}
\toprule
& \multicolumn{5}{c}{$\temp=0$}
& \multicolumn{5}{c}{$\temp=1$} \\
\cmidrule(lr){2-6}\cmidrule(lr){7-11}
& \shortstack{Linear\\$B$=$4$}
& \shortstack{Linear\\$B$=$8$}
& \shortstack{Linear\\$B$=$16$}
& \shortstack{Tree\\$B$=$16$, $V$=$32$}
& \shortstack{Tree\\$B$=$16$, $V$=$60$}
& \shortstack{Linear\\$B$=$4$}
& \shortstack{Linear\\$B$=$8$}
& \shortstack{Linear\\$B$=$16$}
& \shortstack{Tree\\$B$=$16$, $V$=$32$}
& \shortstack{Tree\\$B$=$16$, $V$=$60$} \\
\midrule
\textit{Math} & & & & & & & & & & \\
\quad GSM8K
& 4.14 & 5.52 & 6.28 & 7.02 & 7.27
& 3.99 & 5.14 & 5.67 & 6.29 & 6.49 \\
\quad MATH500
& 4.27 & 5.96 & 7.09 & 7.67 & 8.01
& 4.11 & 5.46 & 6.26 & 6.89 & 7.12 \\
\quad AIME-24
& 4.26 & 5.89 & 6.95 & 7.83 & 8.55
& 4.08 & 5.49 & 6.35 & 6.83 & 6.81 \\
\quad AIME-25
& 4.25 & 5.77 & 6.73 & 7.67 & 8.36
& 4.11 & 5.45 & 6.28 & 6.81 & 7.07 \\
\quad AIME-26
& 4.20 & 6.10 & 7.36 & 7.82 & 8.61
& 4.08 & 5.31 & 6.05 & 6.71 & 6.86 \\
\midrule
\textit{Coding} & & & & & & & & & & \\
\quad HumanEval
& 4.31 & 6.26 & 7.70 & 8.74 & 9.28
& 3.83 & 4.67 & 5.16 & 5.63 & 5.78 \\
\quad MBPP
& 4.42 & 6.55 & 8.34 & 9.16 & 9.45
& 3.87 & 4.86 & 5.20 & 5.87 & 5.96 \\
\quad LCBv6
& 4.29 & 6.03 & 7.09 & 8.16 & 8.20
& 3.77 & 4.51 & 4.74 & 5.34 & 5.53 \\
\midrule
\textit{Science and Knowledge} & & & & & & & & & & \\
\quad GPQA
& 4.35 & 6.07 & 7.15 & 8.16 & 8.52
& 3.79 & 4.64 & 4.91 & 5.49 & 5.65 \\
\quad GPQA-Diamond
& 4.37 & 6.08 & 7.30 & 8.46 & 8.64
& 3.78 & 4.64 & 4.95 & 5.49 & 5.65 \\
\quad MMLU-Pro
& 4.19 & 5.67 & 6.57 & 7.48 & 7.69
& 3.84 & 4.75 & 5.14 & 5.70 & 5.90 \\
\midrule
\textit{Instruction Following} & & & & & & & & & & \\
\quad IFEval
& 4.20 & 5.57 & 6.83 & 7.29 & 7.85
& 3.48 & 4.36 & 4.40 & 4.58 & 4.53 \\
\midrule
Average
& 4.27 & 5.96 & 7.11 & 7.95 & 8.37
& 3.89 & 4.94 & 5.42 & 5.97 & 6.11 \\
\bottomrule
\end{tabular}%
\endgroup
}
\end{subtable}

\medskip

\begin{subtable}{\textwidth}
\centering
\caption{
EAGLE-3 linear and tree samplers.
}
\label{tab:eagle3-acceptance-by-block-unfiltered}
\resizebox{\textwidth}{!}{%
\begingroup
\renewcommand{\arraystretch}{0.85}
\begin{tabular}{l*{14}{c}}
\toprule
& \multicolumn{7}{c}{$\temp=0$}
& \multicolumn{7}{c}{$\temp=1$} \\
\cmidrule(lr){2-8}\cmidrule(lr){9-15}
& \shortstack{Linear\\$B$=$4$}
& \shortstack{Linear\\$B$=$8$}
& \shortstack{Linear\\$B$=$16$}
& \shortstack{Tree\\$B$=$8$, \\$V$=$32$}
& \shortstack{Tree\\$B$=$8$, \\$V$=$60$}
& \shortstack{Tree\\$B$=$16$, \\$V$=$32$}
& \shortstack{Tree\\$B$=$16$, \\$V$=$60$}
& \shortstack{Linear\\$B$=$4$}
& \shortstack{Linear\\$B$=$8$}
& \shortstack{Linear\\$B$=$16$}
& \shortstack{Tree\\$B$=$8$, \\$V$=$32$}
& \shortstack{Tree\\$B$=$8$, \\$V$=$60$}
& \shortstack{Tree\\$B$=$16$, \\$V$=$32$}
& \shortstack{Tree\\$B$=$16$, \\$V$=$60$} \\
\midrule
\textit{Math} & & & & & & & & & & & & & & \\
\quad GSM8K
& 2.43 & 2.68 & 2.73 & 3.82 & 4.07 & 3.78 & 4.24
& 2.26 & 2.48 & 2.49 & 3.62 & 3.83 & 3.57 & 3.83 \\
\quad MATH500
& 2.26 & 2.46 & 2.40 & 3.50 & 3.80 & 3.49 & 3.80
& 2.14 & 2.27 & 2.28 & 3.27 & 3.51 & 3.29 & 3.52 \\
\quad AIME-24
& 2.23 & 2.42 & 2.49 & 3.41 & 3.62 & 3.48 & 3.70
& 2.11 & 2.25 & 2.23 & 3.22 & 3.46 & 3.25 & 3.41 \\
\quad AIME-25
& 2.23 & 2.41 & 2.54 & 3.51 & 3.73 & 3.51 & 3.72
& 2.13 & 2.27 & 2.26 & 3.31 & 3.49 & 3.28 & 3.46 \\
\quad AIME-26
& 2.28 & 2.49 & 2.58 & 3.63 & 3.75 & 3.63 & 3.71
& 2.13 & 2.26 & 2.27 & 3.28 & 3.51 & 3.28 & 3.49 \\
\midrule
\textit{Coding} & & & & & & & & & & & & & & \\
\quad HumanEval
& 2.87 & 3.44 & 3.70 & 4.52 & 5.07 & 4.94 & 5.21
& 2.12 & 2.36 & 2.28 & 3.39 & 3.61 & 3.43 & 3.71 \\
\quad MBPP
& 2.82 & 3.52 & 3.58 & 4.80 & 5.02 & 4.97 & 5.34
& 2.16 & 2.31 & 2.30 & 3.39 & 3.65 & 3.40 & 3.65 \\
\quad LCBv6
& 2.64 & 3.08 & 3.04 & 4.27 & 4.47 & 4.41 & 4.65
& 2.04 & 2.14 & 2.15 & 3.22 & 3.38 & 3.22 & 3.45 \\
\midrule
\textit{Science and Knowledge} & & & & & & & & & & & & & & \\
\quad GPQA
& 2.48 & 2.77 & 2.98 & 3.85 & 4.14 & 4.04 & 4.20
& 1.99 & 2.09 & 2.10 & 3.06 & 3.25 & 3.05 & 3.25 \\
\quad GPQA-Diamond
& 2.49 & 2.77 & 2.93 & 3.82 & 4.17 & 3.87 & 4.09
& 1.98 & 2.07 & 2.07 & 3.00 & 3.19 & 2.99 & 3.19 \\
\quad MMLU-Pro
& 2.35 & 2.60 & 2.62 & 3.68 & 3.93 & 3.69 & 3.95
& 2.03 & 2.14 & 2.14 & 3.11 & 3.27 & 3.10 & 3.31 \\
\midrule
\textit{Instruction Following} & & & & & & & & & & & & & & \\
\quad IFEval
& 2.10 & 3.06 & 3.39 & 3.66 & 3.29 & 4.32 & 4.60
& 1.91 & 2.22 & 2.14 & 3.16 & 3.04 & 3.07 & 3.49 \\
\midrule
$\tps$
& 2.43 & 2.81 & 2.91 & 3.87 & 4.09 & 4.01 & 4.27
& 2.08 & 2.24 & 2.23 & 3.25 & 3.43 & 3.24 & 3.48 \\
\bottomrule
\end{tabular}
\endgroup
}
\end{subtable}

\medskip

\begin{subtable}{\textwidth}
\centering
\caption{Thinking-enabled DFlash at block sizes $B$=$4$, $B$=$8$, and $B$=$16$.}
\label{tab:dflash-acceptance-by-block-unfiltered}
\scalebox{0.9}{%
\begingroup
\renewcommand{\arraystretch}{0.85}
\begin{tabular}{lcccccc}
\toprule
& \multicolumn{3}{c}{$\temp=0$}
& \multicolumn{3}{c}{$\temp=1$} \\
\cmidrule(lr){2-4}\cmidrule(lr){5-7}
& $B$=$4$ & $B$=$8$ & $B$=$16$
& $B$=$4$ & $B$=$8$ & $B$=$16$ \\
\midrule
\textit{Math} & & & & & & \\
\quad GSM8K & $2.54$ & $3.18$ & $3.60$ & $2.47$ & $3.04$ & $3.38$ \\
\quad MATH500 & $2.35$ & $2.99$ & $3.49$ & $2.28$ & $2.82$ & $3.24$ \\
\quad AIME-24 & $2.04$ & $2.48$ & $3.10$ & $1.96$ & $2.22$ & $2.76$ \\
\quad AIME-25 & $2.00$ & $2.30$ & $2.85$ & $1.90$ & $2.23$ & $2.56$ \\
\quad AIME-26 & $1.97$ & $2.37$ & $3.03$ & $1.93$ & $2.19$ & $2.57$ \\
\midrule
\textit{Coding} & & & & & & \\
\quad HumanEval & $2.39$ & $3.12$ & $4.16$ & $2.35$ & $2.71$ & $3.03$ \\
\quad MBPP & $2.28$ & $2.96$ & $3.87$ & $2.22$ & $2.66$ & $3.04$ \\
\quad LCBv6 & $1.92$ & $2.25$ & $2.77$ & $1.76$ & $1.94$ & $2.22$ \\
\midrule
\textit{Science and Knowledge} & & & & & & \\
\quad GPQA & $2.12$ & $2.55$ & $3.23$ & $1.98$ & $2.28$ & $2.57$ \\
\quad GPQA-Diamond & $2.09$ & $2.46$ & $3.13$ & $1.92$ & $2.19$ & $2.50$ \\
\quad MMLU-Pro & $2.23$ & $2.70$ & $3.26$ & $2.11$ & $2.47$ & $2.80$ \\
\midrule
\textit{Instruction Following} & & & & & & \\
\quad IFEval & $2.10$ & $2.71$ & $2.58$ & $1.92$ & $2.23$ & $2.26$ \\
\midrule
$\tps$ & $2.17$ & $2.67$ & $3.26$ & $2.07$ & $2.42$ & $2.74$ \\
\bottomrule
\end{tabular}
\endgroup
}
\end{subtable}
\end{table*}

\begin{table*}[h]
  \centering
  \caption{Median per-request throughput (tok/s/stream) for the 1K-input/8K-output Qwen3-8B workload at $\temp=1$. Each value is the median of three fresh repetitions; ``--'' denotes a resident-capacity-infeasible point.}
  \label{tab:all-methods-throughput}
  \setlength{\tabcolsep}{3.5pt}
  \resizebox{\textwidth}{!}{%
  \begin{tabular}{lrrrrrrrrrrrrrr}
    \toprule
    & \multicolumn{5}{c}{Uno}
    & \multicolumn{5}{c}{EAGLE-3}
    & \multicolumn{3}{c}{DFlash}
    & \multicolumn{1}{c}{AR} \\
    \cmidrule(lr){2-6}\cmidrule(lr){7-11}\cmidrule(lr){12-14}\cmidrule(lr){15-15}
    Con.
    & $B4$ & $B8$ & $B16$ & $B16,V32$ & $B16,V60$
    & $B4$ & $B8$ & $B16$ & $B16,V32$ & $B16,V60$
    & $B4$ & $B8$ & $B16$
    & -- \\
    \midrule
    \multicolumn{15}{c}{{System throughput}} \\
    \midrule
    1   & 305  & 366  & 416  & 445  & 435  & 204  & 198  & 161  & 289  & 284  & 287  & 331  & 369  & 176 \\
    2   & 566  & 703  & 768  & 803  & 730  & 401  & 377  & 312  & 531  & 504  & 538  & 645  & 702  & 352 \\
    4   & 1086 & 1205 & 1420 & 1411 & 1221 & 784  & 737  & 585  & 906  & 771  & 1058 & 1188 & 1296 & 653 \\
    8   & 1906 & 2397 & 2384 & 2161 & 1630 & 1381 & 1292 & 1038 & 1399 & 1019 & 1874 & 2073 & 2153 & 1144 \\
    16  & 3247 & 3814 & 3436 & 2798 & 1869 & 2429 & 2233 & 1693 & 1746 & 1118 & 3075 & 3315 & 3007 & 1933 \\
    32  & 4586 & 4873 & 4086 & 2958 & 1964 & 3745 & 3308 & 2228 & 1881 & 1171 & 4407 & 4246 & 3396 & 2796 \\
    64  & 5733 & 5600 & 4112 & 3138 & 2004 & 4944 & 3965 & 2580 & 1969 & --     & 5351 & 4900 & 3592 & 3662 \\
    128 & --     & --     & --     & --     & --     & --     & --     & --     & --     & --     & --     & --     & --     & -- \\
    \midrule
    \multicolumn{15}{c}{{Per-user throughput (tokens/s/user)}} \\
    \midrule
    1   & 305 & 366 & 416 & 445 & 435 & 204 & 198 & 161 & 289 & 284 & 287 & 331 & 369 & 176 \\
    2   & 283 & 351 & 384 & 401 & 365 & 200 & 189 & 156 & 266 & 252 & 269 & 322 & 351 & 176 \\
    4   & 271 & 301 & 355 & 353 & 305 & 196 & 184 & 146 & 227 & 193 & 264 & 297 & 324 & 163 \\
    8   & 238 & 300 & 298 & 270 & 204 & 173 & 161 & 130 & 175 & 127 & 234 & 259 & 269 & 143 \\
    16  & 203 & 238 & 215 & 175 & 117 & 152 & 140 & 106 & 109 & 70  & 192 & 207 & 188 & 121 \\
    32  & 143 & 152 & 128 & 92  & 61  & 117 & 103 & 70  & 59  & 37  & 138 & 133 & 106 & 87 \\
    64  & 90  & 87  & 64  & 49  & 31  & 77  & 62  & 40  & 31  & --    & 84  & 77  & 56  & 57 \\
    128 & --    & --    & --    & --    & --    & --    & --    & --    & --    & --    & --    & --    & --    & -- \\
    \bottomrule
  \end{tabular}%
  }
\end{table*}

\end{document}